\documentclass[preprint,12pt]{elsarticle}

\usepackage[T1]{fontenc}
\usepackage{graphicx,verbatim}

\usepackage{booktabs}
\usepackage{multirow}

\usepackage{amsfonts}
\usepackage{amssymb}
\usepackage{amsmath}
\usepackage{subfig}
\usepackage{mathtools}
\usepackage{hyperref}
\usepackage{pifont}

\begin{document}

\begin{frontmatter}



\title{Interpretability and Individuality in Knee MRI: Patient-Specific Radiomic Fingerprint with Reconstructed Healthy Personas}


\author[label1,label2]{Yaxi Chen}
\author[label3,label4]{Simin Ni}
\author[label1]{Shuai Li}
\author[label2,label5,label6,label7]{Shaheer U. Saeed}
\author[label3,label4]{Aleksandra Ivanova}
\author[label4]{Rikin Hargunani}
\author[label1]{Jie Huang}
\author[label3,label4]{Chaozong Liu}
\author[label2,label7]{Yipeng Hu}

\affiliation[label1]{%
    organization={Department of Mechanical Engineering, University College London},
    city={London},
    country={UK}
}

\affiliation[label2]{%
    organization={Hawkes Institute, University College London},
    city={London},
    country={UK}
}

\affiliation[label3]{%
    organization={Institute of Orthopaedic \& Musculoskeletal Science, University College London, Royal National Orthopaedic Hospital},
    city={Stanmore},
    country={UK}
}

\affiliation[label4]{%
    organization={Royal National Orthopaedic Hospital},
    city={Stanmore},
    country={UK}
}
\affiliation[label5]{%
    organization={School of Engineering and Materials Science, Queen Mary University of London},
    city={London},
    country={UK}
}

\affiliation[label6]{%
    organization={Centre for Bioengineering, Queen Mary University of London},
    city={London},
    country={UK}
}

\affiliation[label7]{%
    organization={Department of Medical Physics and Biomedical Engineering, University College London},
    city={London},
    country={UK}
}

\begin{abstract}
For automated assessment of knee magnetic resonance imaging (MRI) scans, both predictive accuracy and interpretability are essential for clinical use and its responsible adoption. Traditional radiomic models rely on predefined radiomic feature chosen at the population level, which, while much more interpretable, are too restrictive to capture highly variable patient-specific characteristics (that is useful for discriminative tasks) and often underperform compared to recent end-to-end deep learning (DL), in our classification applications. To address this limitation, we propose for this application two complementary strategies that bring patient-level individuality and interpretability: radiomic fingerprints and healthy personas.  First, a radiomic fingerprint is a dynamically constructed, patient-specific set of features derived from MRI.  Instead of applying a uniform population-level signature, our model predicts feature relevance from a large pool of candidate radiomic features and selects only those most predictive for individual patients, while maintaining feature-level interpretability. Our radiomic fingerprint can be viewed as a latent-variable model of patient-specific feature usage, where an image-conditioned predictor estimates usage probabilities and a transparent logistic regression with global coefficients performs classification.
Second, a healthy persona adds to interpretability by synthesizing a pathology-free baseline for each patient using a denoising diffusion probabilistic model trained for reconstructing healthy knee MRIs. Comparing features extracted from pathological images against their reconstructed personas highlights potential deviations from normal anatomy, enabling intuitive, case-specific explanations of disease manifestations. It is in addition to the computational advantage due to the resulting patient-level residual learning. We systematically compare the efficiency and efficacy of the proposed fingerprints, personas and their combination, across three clinical tasks: detecting general abnormalities, anterior cruciate ligament (ACL) tears and meniscus tears. Experimental results show that both approaches yield performance comparable to or surpassing state-of-the-art DL models, while uniquely supporting interpretability at multiple aspects: localised feature-level transparency via fingerprints and patient-level explanations via personas. Case studies further illustrate how these perspectives facilitate potential human-explainable biomarker discovery and pathology localisation. Our findings demonstrate that combining patient-specific fingerprints with generative healthy personas offers a powerful and interpretable alternative to black-box DL, advancing radiomic analysis for clinical decision support. Code is available at: \url{https://github.com/YaxiiC/RadiomicsPersona.git}

\end{abstract}




\end{frontmatter}



\section{Introduction}
\label{sec:intro} 

\textbf{Clinical motivation:} Anterior cruciate ligament (ACL) and meniscus injuries are among the most common musculoskeletal disorders, significantly affecting mobility and quality of life~\cite{salzler2015state}. These injuries often necessitate timely and accurate diagnosis to guide appropriate treatment and rehabilitation. While physical examination techniques, such as the Lachman and McMurray tests, offer moderate sensitivity and specificity for detecting ACL and meniscal injuries, ranging from 67\% to 96\% ~\cite{leblanc2015diagnostic, kim2008effect}. 
Magnetic resonance imaging (MRI) enables accurate identification of ligament integrity, meniscal morphology, and associated joint pathology, thereby serving as the primary imaging tool for differential diagnosis~\cite{blyth2015diagnostic}. This detailed anatomical and pathological information is essential for guiding clinical decision-making, including the choice between conservative management and surgical intervention, and for tailoring the surgical approach when operative treatment is indicated. However, knee MRI examinations generate a high volume of detailed image slices, requiring substantial reading time for radiologists, and even experienced musculoskeletal specialists are susceptible to inter-observer variability in interpretation ~\cite{kim2008effect}. For example, Krampla et al. had 11 radiologists independently read 52 knee MRIs and reported low inter-observer agreement, with values of 0.37 for cartilage lesions and 0.60 for meniscal tears~\cite{krampla2009mri}. The development of automated systems for interpreting knee MRI images is a promising area. Such automated interpretation systems could provide essential diagnostic information, including tear depth, location, pattern, tissue quality, tear length, and details regarding previous meniscal interventions or repairs~\cite{hegedus2007physical}, thereby potentially improving diagnostic accuracy and clinical efficiency. For clinical adoption high accuracy alone is not sufficient, the systems must also provide interpretable outputs that allow clinicians to verify where the model is focusing and which image features drive its decision ~\cite{jin2023guidelines, borys2023explainable}. In practice, this means producing spatially meaningful evidence (e.g., localised cues around the suspected tear) that aligns with expert expectations and can be evaluated for plausibility during routine reading.

\noindent\textbf{Deep learning (DL) for knee MRI: }For disease classification, end-to-end DL algorithms can efficiently identify and categorize various knee pathologies, reducing radiologists' workloads and allowing them to concentrate on more subtle or challenging cases~\cite{bien2018deep}. Additionally, DL methods have been successfully employed in predicting disease stages, utilizing longitudinal data to anticipate disease progression, which is particularly valuable for conditions like osteoarthritis. Such predictive algorithms offer practical value by forecasting disease onset in healthy individuals, enabling early interventions. Furthermore, anomaly detection using DL, such as the automatic segmentation of cartilage and menisci followed by classification of structural abnormalities, significantly enhances diagnostic accuracy and reliability~\cite{liu2024aclnet,chen2024segmentation}. The integration of these DL techniques promises substantial improvements in clinical workflow efficiency, diagnostic consistency, and overall patient care outcomes. However, despite these advances, the widespread clinical adoption of DL models is limited by their “black-box” nature. The decision-making processes of these models are not transparent, which poses a significant challenge, especially in high-risk medical settings~\cite{marey2024explainability}. 


\noindent\textbf{Interpretability-radiomics: }One established branch of interpretability in medical imaging is radiomics, which has long contributed to screening, diagnosis, treatment planning, and follow-up~\cite{zhang2023radiomics}. By extracting quantitative features from medical images in a reproducible manner, radiomics enables the development of robust predictive models that correlate these features with various clinical outcomes. Hand-crafted features, such as intensity, shape, texture and wavelets, provide predefined and more understandable information regarding the region of interest (ROI)~\cite{soffer2019convolutional}. Features are extracted using well-defined mathematical operators applied directly to the image data, producing standardized quantitative descriptors of the ROI. This process offers a consistent and reproducible characterization of image patterns, complementing conventional visual assessment with measurable evidence ~\cite{ardakani2022interpretation}. 

Although the majority of radiomics studies have focused on oncology applications, its potential extends beyond cancer imaging. Radiomics can offer a quantitative signature of tumor characteristics that are not readily appreciated visually, showing promise in identifying tumor subtypes and predicting outcomes by integrating radiomic features with clinical variables via machine learning methods. Moreover, recent applications have begun to explore radiomics in the discovery of knee osteoarthritis imaging features for both diagnosis and prognosis~\cite{jiang2024radiomics}, further highlighting the versatility of this method. However, challenges remain: radiomic features can be sensitive to acquisition variability, require robust feature selection, and may be less expressive than deep representations.

\noindent\textbf{Interpretability-visual explanations: }A complementary dimension of interpretability emerges from visual explanations, which aim to localize and contextualize the image regions driving model predictions. Class Activation Maps (CAM) and their variants, such as Grad-CAM~\cite{selvaraju2016grad}, have been widely applied in medical imaging to highlight salient areas within pathological regions~\cite{guo2024liver}. For example, Yang et al.~\cite{yang2025development} demonstrated a CAM-guided framework for predicting lymph node metastasis from multiparametric MRI in rectal cancer, illustrating how heatmaps can provide intuitive cues regarding model focus. However, CAM-based methods have notable drawbacks: they depend heavily on the final convolutional layer, are sensitive to noise, and cannot guarantee that the highlighted areas correspond to genuine lesion-related information. In parallel, advances in generative modelling have introduced new opportunities for visual interpretability. Generative models have gained significant attention in medical imaging for their ability to synthesise realistic, high-resolution data. A range of powerful deep generative architectures has been developed to model complex data distributions, including diffusion models~\cite{ho2020denoising, rombach2022high}, autoregressive transformers~\cite{parmar2018image, esser2021taming}, generative adversarial networks (GANs) ~\cite{goodfellow2020generative} and variational autoencoders (VAEs) ~\cite{kingma2013auto}. Among these, Denoising Diffusion Probabilistic Models (DDPMs) have demonstrated superior performance in producing anatomically plausible images by learning to reverse the gradual corruption of data through a diffusion process ~\cite{Khader2023Denoising}. In clinical contexts, generative models offer unique advantages, such as creating synthetic training data, simulating missing or corrupted anatomy, and enhancing interpretability through counterfactual examples. By synthesizing a healthy baseline, generative approaches enable clinicians to directly visualize how patient anatomy deviates from a hypothetical disease-free state, thereby complementing radiomics with intuitive, image-level interpretability.  

\begin{figure}
    \centering
     \includegraphics[width=0.9\textwidth]{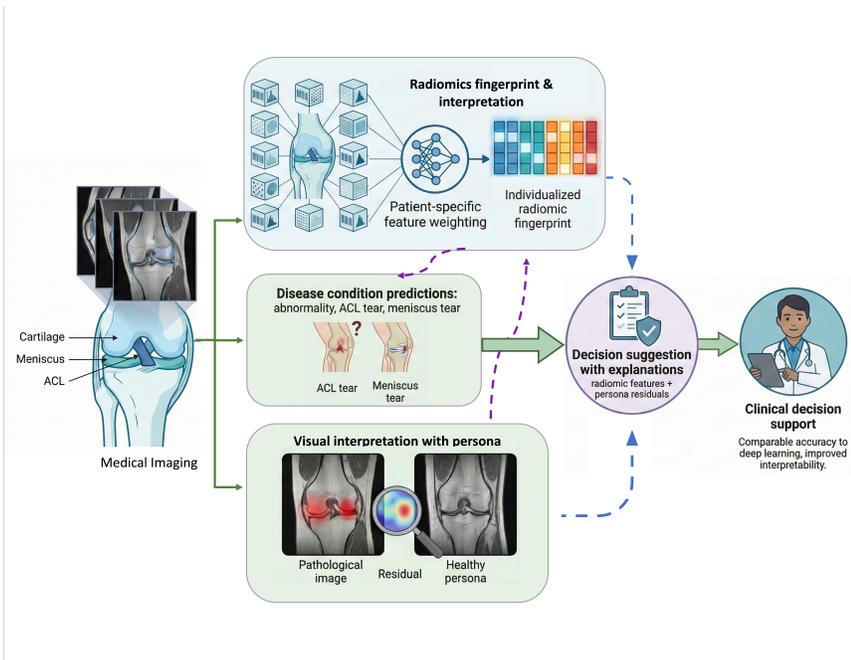}
    \caption{Graphical abstract. Overview of the proposed framework, patient-specific radiomic fingerprints with reconstructed healthy personas.} 
    \label{graphical_abstract}
\end{figure}

In this study, we present a framework that integrates two complementary sources of interpretability: quantitative radiomic descriptors and generatively reconstructed healthy personas, enabling more transparent disease assessment. Our approach employs a probabilistic feature-weighted classification framework in which a usage predictor estimates subject-specific feature usage probabilities over a large set of candidate radiomic features extracted from candidate ROIs. These usage weights modulate how each candidate radiomic feature contributes to a transparent downstream classifier for the clinical task. The usage predictor and the downstream classifier are trained jointly using only the downstream task labels. 
We further present a probabilistic interpretation as a latent-variable model with binary usage indicators marginalised for prediction, providing a practical continuous relaxation. To enhance the clinical relevance and interpretability of this feature usage process, we introduce the concept of a healthy persona, a synthesized, subject-specific healthy baseline for pathological regions. Inspired by the psychological notion of persona \cite{jung2014two}, this construct represents an individualized, pathology-free version of the patient's anatomy, enabling direct comparisons between pathological and healthy tissue. We generate the healthy persona using a DDPM trained exclusively on healthy MRI scans. The model reconstructs pathology-free versions of ROIs, effectively simulating what the patient’s anatomy would look like without the disease. By integrating the patient-specific healthy persona with subject-specific feature usage probabilities over candidate radiomic features, our framework delivers a clinically grounded and interpretable approach to musculoskeletal disease assessment.

To summarize, our main contributions are:
\begin{enumerate}
\item We introduce a probabilistic feature-weighted framework that learns subject-specific feature usage probabilities, enabling more precise disease characterization and interpretation;
\item We provide a latent-variable, probabilistic interpretation of the proposed patient-specific radiomic fingerprint, clarifying the roles of feature usage predictor and classifier, and their relationships to classical radiomics and end-to-end DL.
\item We propose a method to reconstruct a healthy persona for each patient’s pathological image to provide an ROI-wise healthy baseline for direct comparison;
\item We achieved comparable or even superior performance compared to traditional end-to-end approaches in knee joint MRI analysis, while offering enhanced interpretability, flexibility and revealing potential imaging biomarkers.
\end{enumerate}

This study builds upon and substantially extends our preliminary work presented at MICCAI 2025~\cite{chen2025patient}. Relative to the conference version, this journal submission expands the scope (theoretical and empirical), increases methodological rigour and strengthens reproducibility.

We substantially expand the Methods to formalise the approach as an explicit latent-variable model. In particular, subject-specific latent feature-usage indicators $\mathbf{z}$ (or their continuous relaxation $\mathbf{u}$) are inferred from the input image, while a transparent logistic regression with global coefficients $\boldsymbol{\beta}$ performs classification conditional on feature usage. This formulation clarifies the separation between population-level effects ($\boldsymbol{\beta}$) and individualised feature reliance ($\mathbf{u}$), and recasts the learning objective as a practical relaxation of the marginalised latent model. Additional extensions include:

\textit{\textbf{Broader baselines under matched settings:}} a consistently configured set of comparators spanning classical radiomics pipelines (feature extraction, feature selection and conventional classifiers) and end-to-end deep learning models trained under matched conditions.

\textit{\textbf{Reproducibility and implementation details:}} complete reporting of architectures, hyperparameters, training schedules and evaluation protocol, together with an open-source repository to reproduce all experiments and figures.

\textit{\textbf{Expanded generative modelling evaluation:} }assessment of additional generative models for healthy-persona reconstruction, including quantitative reconstruction fidelity metrics and qualitative examples.

\textit{\textbf{Additional datasets and generalisation analysis:}} expanded experiments on additional knee MRI data, including explicit cross-dataset generalisation analyses to probe robustness under dataset shift.

\textit{\textbf{Larger radiomic feature pool and ablations:}} extension of the radiomic feature set (including higher-order texture families), with ablation studies to quantify interpretability--performance trade-offs.

\textit{\textbf{Consistent probabilistic formalisation:}} explicit formulation of the proposed method as a latent-variable probabilistic model of subject-specific feature usage, including notation, marginalisation and a continuous relaxation that clarifies links to classical radiomics and end-to-end DL.

These expanded discussion, methodologies, implementation details, data sets and experiments enhance the generalisability, robustness and clinical applicability of the proposed approach.

\section{Related Work}

\subsection{End-to-end approaches for knee MRI analysis}
DL has been employed in most recent automated approaches to MR-based diagnosis of knee injuries. Most previous studies in this area have utilized end-to-end CNNs. For example, Bien et al.~\cite{bien2018deep} utilized an end-to-end classifier to detect abnormalities, ACL tears and meniscal tears from knee MRI, demonstrating performance comparable to that of experienced radiologists. Tsai et al.~\cite{tsai2020knee} introduced ELNet, a lightweight and efficiently layered CNN tailored for computational efficiency while maintaining diagnostic accuracy.

While CAM–based methods~\cite{bien2018deep} offer useful spatial interpretability by highlighting image regions that contribute to a model’s prediction. These methods do not reveal which underlying imaging characteristics, such as intensity patterns, texture, morphology, or other feature-level cues, actually drive the decision. Furthermore, end-to-end models can be susceptible to dataset biases and may fail to generalize across institutions or imaging protocols without extensive retraining or adaptation~\cite{zech2018variable}.

\subsection{DL in radiomic analysis of musculoskeletal imaging}

Radiomics and DL have traditionally been treated as separate paradigms in medical image analysis, radiomics offering interpretability through handcrafted features, and DL delivering high performance through automated feature learning. However, recent studies have increasingly explored hybrid approaches to leverage the strengths of both, for example, in tumor characterization, to improve diagnostic performance while maintaining a degree of interpretability essential for clinical adoption~\cite{fritz2023radiomics}. A recent study developed a DL radiomics model based on spinal X-ray images to predict and differentiate acute from chronic osteoporotic vertebral fractures, achieving high diagnostic accuracy through extracting DL features and radiomic features in separate branches and then fusing them at a later stage~\cite{zhang2024constructing}. Moreover, the review by Das et al.~\cite{das2023review} highlighted a shift towards hybrid DL–radiomics pipelines in musculoskeletal imaging, in which deep networks are used to learn image representations that are then combined with hand-crafted radiomic features and conventional machine-learning classifiers. Such hybrid strategies have been explored for tasks including tumour characterisation, osteoarthritis-severity grading and fracture-risk prediction.

\subsection{Generative modelling for patient-specific baselines and analysis} 

GANs, VAEs and diffusion models have become increasingly valuable in medical imaging for tasks including data augmentation, reconstruction, anomaly detection and image-to-image translation~\cite{kazerouni2022diffusion, Khader2023Denoising}. However, their use in diagnostic workflows remains relatively underexplored, particularly for producing diagnostically meaningful images that improve interpretability and support counterfactual analysis. To illustrate existing progress in this area, several generative-model applications have been reported across different imaging modalities. In brain MRI, GEM-3D~\cite{zhu2024generative} employs a conditional diffusion framework to generate volumetric 3D medical images from partial input, simulating individualised healthy anatomy. Likewise, Kumar et al.~\cite{kumar2022counterfactual} introduced a conditional generative model that perturbs local image features to synthesize counterfactuals representing alternative disease outcomes, thereby aiding in the discovery of personalised predictive imaging markers. A comprehensive review~\cite{ibrahim2025generative} also emphasizes how generative models are being adopted across modalities to simulate diverse, patient-matched imaging data for clinical and research use.
In musculoskeletal imaging, Colussi et al.~\cite{colussi2024loris} proposed a localised reconstruction framework for lesion detection and segmentation using a generative inpainting model. Their approach effectively synthesizes healthy tissue appearances in localised masked regions, allowing for direct comparison between pathological and estimated normal anatomy, thus facilitating counterfactual reasoning.
Despite these advances, the use of generative models for constructing patient-specific baselines and generating counterfactual examples remains relatively underexplored. This direction is particularly valuable in clinical settings where interpretability and individualized assessment are essential.

\section{Preliminary: Classical Radiomic Classification v.s. End-to-End DL}

In this section, we describe three types of approaches previously used in the field of medical image analysis for this application: (1) a classical radiomics approach, (2) hybrid approaches that integrate DL and (3) end-to-end DL. This clarifies the differences between the proposed methods (Sec.~\ref{sec:method}) and these existing approaches using consistent terminologies and notations.

\subsection{Classical radiomic feature classification}
\label{sec:preliminary_traditional_radiomics}

In classical radiomics, predictive modelling typically follows a two-stage process: (1) extracting a set of predefined candidate radiomic features, from one or more ROIs, and (2) classifying~\footnote{In this paper, we discuss with a binary classification as an example, which may be extended to multiclass classification or regression tasks.} the condition(s) of interest via supervised learning. 

Assume a set of $J$ ROIs $\{\mathbf{x}_{j}\}$ for a given medical image $\mathcal{I}$, where $j=1,\dots,J$. In general, such ROIs can be defined by segmentation masks of the same full image size \cite{van2017computational}. Now, a column vector containing $N$ radiomic features $r_n$, where $n=1,\dots,N$, which can be computed for this image:

\begin{equation}
\label{equ:radiomic_feature_vector}
\mathbf{r} = [r_{1}, r_{2}, \dots, r_{N}]^\top = [{\mathbf{r}^{(1)}}^\top, {\mathbf{r}^{(2)}}^\top, \dots, {\mathbf{r}^{(J)}}^\top,]^\top
\end{equation}
where $\mathbf{r}^{(j)} = e^{(j)}(\mathbf{x}_j)$ contains $K_j$ radiomic features, extracted from the $j^{th}$ ROI by the feature extracting function $e^{(j)}(\cdot)$, and $N=\sum_{j=1}^{J}K_j$.

In a typical radiomic feature analysis, 1) the ROIs $\{\mathbf{x}_{j}\}_{j=1}^J$ and 2) the feature extracting functions $e^{(j)}(\cdot)$ (hence the extracted features $\mathbf{r}^{(j)}$) are both predefined. In fact, the ROI definitions and the search for the relevant types of features have been focus of radiomic research \cite{lambin2017radiomics}. 

Once the radiomic feature vector $\mathbf{r}\in \mathbb{R}^{N}$ is extracted, a parametric classifier $f_{\boldsymbol{\theta}}(\mathbf{r})$ with parameter vector $\boldsymbol{\theta}$ is fitted to estimate class probabilities $p(\mathbf{c} \mid \cdot)$ of a binary condition of interest $\mathbf{c} \in \{1,2\}$, $1$ being positive (e.g. presence of disease or pathology) and $\sum_{\mathbf{c}}p(\mathbf{c}\mid \cdot)=1$:
\begin{equation}
\label{eq_prob_radiomics}
p(\mathbf{c}=1 \mid \mathcal{I})=p(\mathbf{c}=1 \mid \boldsymbol{\theta})=f_{\boldsymbol{\theta}}(\mathbf{r})=f_{\boldsymbol{\theta}}(f^{(\textit{rad})}(\mathcal{I}))
\end{equation}
where 
\begin{equation}
\label{eq_radiomic_extraction}
\mathbf{r}=f^{(\textit{rad})}(\mathcal{I})
\end{equation}
denotes the process of the aforementioned ROI-based radiomic extraction  $e^{(j)}(\mathbf{x}_j)$ from the given image $\mathcal{I}$, and $f^{(\textit{rad})}$ has no learnable parameters during training the classifier.
To maintain intepretability of the classification system, low-dimensional, linear, monotonic or other easy-to-explain functions are usually required for $f_{\boldsymbol{\theta}}(\cdot)$, such as low-degree polynomial functions, logistic regression and support vector machines, exemplified in existing work discussed in Sec.~\ref{sec:intro}.

\subsection{Hybrid approaches that integrate DL}
\label{sec:preliminary_hybrid}

\paragraph{\textbf{DL for automating ROI delineation}}
Deep neural networks can be used to automatically delineate ROIs, after which classical radiomic features are used for classification as described in Sec.~\ref{sec:preliminary_traditional_radiomics}. Segmentation is arguably one of the most successful applications of DL in medical image analysis, with a high volume of existing literature \cite{ronneberger2015u, isensee2021nnu, litjens2017survey, rayed2024deep}. 

It is noteworthy that such an approach still requires the types of ROIs being predefined, before training a separate (thus fixed) segmentation network $s^{(j)}(\mathcal{I})$ for $j^{th}$ ROI. Whilst the ROI delineation and radiomic feature extraction process (Eq.~\ref{eq_radiomic_extraction}) is now fully automated, the probabilistic view remains the same as in Eq.~\ref{eq_prob_radiomics}, where the classifer parameters $\boldsymbol{\theta}$ remain the leanrable parameters that the class probabilities are conditioned on: 
\begin{equation}
\label{eq_prob_radiomics_seg}
p(\mathbf{c}=1 \mid \mathcal{I})=p(\mathbf{c}=1 \mid \boldsymbol{\theta})=f_{\boldsymbol{\theta}}(\mathbf{r})=f_{\boldsymbol{\theta}}(f^{(\textit{rad*})}(\mathcal{I}))
\end{equation}
where
$\mathbf{r}=f^{(\textit{rad*})}(\mathcal{I})=[{\mathbf{r}^{(1)}}^\top, {\mathbf{r}^{(2)}}^\top, \dots, {\mathbf{r}^{(J)}}^\top,]^\top$ and $\mathbf{r}^{(j)} = e^{(j)}(s^{(j)}(\mathcal{I}))$.

Interestingly, one could consider to oppitmise segmentation network $s^{(j)}(\mathcal{I})$, together with the classifier parameters $\boldsymbol{\theta}$ without predefined ROI definitions, which turns a partial, nontrivial weakly-supervised segmentation (supervised by labels of condition classes $\mathbf{c}$). This formulation is related to the proposed methods and will be discussed in Sec.~\ref{sec:method}.

\paragraph{\textbf{DL for classifying ROI-based radiomic features}}
Deep neural networks can simply replace the classifer $f_{\boldsymbol{\theta}}$ in Eq.~\ref{eq_prob_radiomics}, in which $\boldsymbol{\theta}$ becomes the network weights. However, the deep neural networks lose the intepretability the classical classifiers offer \cite{papadimitroulas2021artificial, parekh2019deep}, and may be considered equivalent to feature weighing or selecting using DL models. Their reduced explainability is discussed further in Sec.~\ref{sec:method}. With unclear advantage over end-to-end learning (Sec.~\ref{sec:end_to_end}), this may partly explain why this approach has not been widely adopted.

In our knee MR imaging application, there has been little evidence that suggested classifers in classical radiomics lacked representation or generalisation abilities for their intended classification tasks - mapping from lower-dimensional radiomic feature vectors to condition class probabilities (i.e., $p(\mathbf{c})=f_{\boldsymbol{\theta}}(\mathbf{r}$)). We argue that, in contrast, underfitting or a lack of strong correlation between radiomic features and conditions of interest has challenged the definition of (and the search for) better, more condition-predictive feature types, including the ROI definitions (Eq.~\ref{eq_radiomic_extraction}).

\subsection{End-to-end DL image classification}
\label{sec:end_to_end}

In contrast, end-to-end approaches do not explicitly involve ROI-based radiomic features:
\begin{equation}
\label{eq_prob_end2end}
p(\mathbf{c}=1 \mid \mathcal{I})=p(\mathbf{c}=1 \mid \boldsymbol{\phi}) = f_{\boldsymbol{\phi}}(\mathcal{I})
\end{equation}
where $f_{\boldsymbol{\phi}}(\mathcal{I})$ is a neural network with weights $\boldsymbol{\phi}$, taking image $\mathcal{I}$ as input. The network weights $\boldsymbol{\phi}$ become learnable parameters during training of the classification. It has however been well-documented that the representation learning using over-prameterised $f_{\boldsymbol{\phi}}$ often leads to high-dimensional latent features that are difficult to interpret or to align with human intuition \cite{montavon2018methods, samek2021explaining, salahuddin2022transparency}.

\section{Proposed Method}
\label{sec:method}

\subsection{Candidate radiomic features}
Following the notations used in Sec.~\ref{sec:preliminary_traditional_radiomics}--\ref{sec:end_to_end}, consider a set of $K_j$ candidate radiomic features extracted from each of $J$ candidate ROIs, resulting in a total of $N=\sum_{j=1}^{J}K_j$ candidate radiomic features.

First, the candidate radiomic features are not tailored or manually selected for the specific application. Instead, we include a comprehensive set of commonly used radiomic descriptors, including 3D shape features and first-order statistical features, computed in a standardised manner.

Second, the candidate ROIs are sampled from the given image $\mathcal{I}$. In principle, these ROIs can be sampled from the entire image volume, e.g., using exhaustive image patches or moving windows. In practice, however, the number of candidate ROIs can be substantially reduced, for example by restricting the ROI sampling to anatomically relevant regions or a spatial neighbourhood of interest - this is hereinafter referred to as ROI sampling domain. In this study, we investigate different configurations for the ROI sampling domain, including $2\times2\times2$ and $3\times3\times3$ regular 3D patches partitioning the ROI sampling domain, as detailed in Sec.~\ref{sec:Feature Extraction}.

The combination of multiple candidate ROIs and multiple radiomic feature types yields a large but explicit set of $N$ candidate radiomic features, which forms the input to the proposed probabilistic feature-weighted classification framework described next.

\subsection{Individualised probabilistic feature selection}
\label{sec:method_individualised}

Classical radiomics relies on predefined ROI definitions and radiomic feature types (Sec.~\ref{sec:preliminary_traditional_radiomics}). 
We propose not only to automate these choices, but to learn subject-specific feature usage probabilities, while retaining an explicit radiomic decision form for interpretability.

Assume a neural network $f_{\boldsymbol{\alpha}}(\cdot)$ with parameters $\boldsymbol{\alpha}$ that predicts a vector of feature ``usage weights'' $r_n$ from the input image:

\begin{equation}
\label{eq:u_predictor}
\mathbf{u} = f_{\boldsymbol{\alpha}}(\mathcal{I}) \in [0,1]^N,
\end{equation}
where $\mathbf{u} = [u_1, u_2, ..., u_N]^\top$ and the $u_n$ is range-enforced, for example, by an element-wise sigmoid activation at the network output.
Given $u_n$ and the candidate radiomic features $r_n$, the condition class probability is modelled via a logistic regression classifier:

\begin{equation}
\label{eq:main_model}
p(\mathbf{c}=1 \mid \mathcal{I}) = p(\mathbf{c}=1 \mid \boldsymbol{\alpha},\boldsymbol{\beta}) = \sigma\!\left(
\sum_{n=1}^{N} \beta_n \, u_n \, r_n \right),
\end{equation}
where $\sigma(\cdot)$ is the logistic sigmoid, and $\beta_n$ are the logistic regression coefficients and $\boldsymbol{\beta}=[\beta_1, \beta_2, ..., \beta_N]^\top$. 

It is important to clarify the roles of $\boldsymbol{\beta}$ and $\mathbf{u}$. In Eq.~\ref{eq:main_model}, $\beta_n$ and $u_n$ appear as a product, but they have distinct and complementary interpretations. The coefficient $\beta_n$ are fixed parameters (once the classifier is trained) shared across the dataset. It quantifies the direction and magnitude of association between radiomic feature $r_n$ and condition $\mathbf{c}$, i.e., how strongly $r_n$ contributes to the prediction, when it is used. In contrast, $u_n$ are image-dependent and predicted by the network $f_{\boldsymbol{\alpha}}$. It varies across subjects and represents the probability that feature $r_n$ is relied upon for a particular image.

Thus, $\boldsymbol{\beta}$ explains what each radiomic feature means for prediction at the population level, while $\mathbf{u}$ explains which features are used for an individual subject. Collapsing them into a single parameter would lose this separation between global effect size and subject-specific feature usage, and would reduce the interpretability of the model. 

\subsection{Joint training of feature usage and downstream classification}
\label{sec:method_training}

The parameters $\boldsymbol{\alpha}$ (usage predictor) and $\boldsymbol{\beta}$ (logistic regression) are trained jointly using only condition labels $\mathbf{c}$. 
Let $(\mathcal{I},\mathbf{c}) \sim p_{\text{data}}$ denote samples from the training distribution. The expected negative log-likelihood is minimised:
\begin{equation}
\label{eq:training_objective_expectation}
\min_{\boldsymbol{\alpha},\boldsymbol{\beta}} 
\;\mathbb{E}_{(\mathcal{I},\mathbf{c}) \sim p_{\text{data}}}
\Big[
\mathcal{L}\!\Big(\mathbf{c},\;
p(\mathbf{c} \mid \boldsymbol{\alpha},\boldsymbol{\beta})\Big)
\;+\;\lambda_{u}\,\|\mathbf{u}\|_{1}
\Big]
\;+\;\lambda_{\beta}\|\boldsymbol{\beta}\|_{2},
\end{equation}
where $\mathcal{L}(\cdot,\cdot)$ thus is the cross-entropy between condition labels and predicted probabilities. The $L^1$ and $L^2$ regularisers, on $\mathbf{u}$ and $\boldsymbol{\beta}$, encourage sparsity for feature usage at subject level and model incomplexity at the population level, and are weighted by two hyperparameters $\lambda_{u}$ and $\lambda_{\beta}$, respectively.

The full objective is end-to-end differentiable in both $\boldsymbol{\alpha}$ and $\boldsymbol{\beta}$ (Eq.~\ref{eq:main_model}), enabling standard gradient-based optimisation. Once trained, the predicted usage probabilities $\mathbf{u}$ can optionally be binarised by thresholding to yield a hard feature-selection mask, so that only a small subset of radiomic features is used per subject, further enhancing interpretability.

\subsection{Probabilistic interpretation and relation to existing approaches}
\label{sec:method_probabilistic_view}

The proposed formulation provides a clear probabilistic interpretation as image-conditioned feature usage/selection, based on an explicit radiomic representation. 
Assume latent binary usage variables $\mathbf{z}=[z_1,\dots,z_N]^\top\in\{0,1\}^N$, where $z_n=1$ indicates that feature $r_n$ is used for prediction for a given subject/image.
We model $\mathbf{z}$ as independent Bernoulli random variables, $p(\mathbf{z}\mid \cdot)=\prod_{n=1}^{N} u_n^{z_n}\big(1-u_n\big)^{1-z_n}$. Its probability are estimated by neural network $f_{\boldsymbol{\alpha}}$ from the input image via Eq.~\ref{eq:u_predictor}, thus the conditional probability:
\begin{equation}
\label{eq:bern_usage}
u_n = p(z_n=1\mid \mathcal{I},\boldsymbol{\alpha}),
\end{equation}
In turn, final class prediction is given by the logistic regression classifier, conditioned on the feature usage $\mathbf{z}$ and radiomic features $\mathbf{r}$:
\begin{equation}
\label{eq:cond_on_z}
p(\mathbf{c}=1 \mid \mathbf{r},\mathbf{z},\boldsymbol{\beta})
=\sigma\!\left(\sum_{n=1}^{N}\beta_n z_n r_n\right)
\end{equation}
Marginalising out $\mathbf{z}$, we have:
\begin{equation}
\label{eq:marginal_z}
p(\mathbf{c}\mid \mathcal{I})
=\sum_{\mathbf{z}}
p(\mathbf{c}\mid \mathbf{r},\mathbf{z},\boldsymbol{\beta})\;
p(\mathbf{z}\mid \mathcal{I},\boldsymbol{\alpha}).
\end{equation}
Eq.~\ref{eq:main_model} can be viewed as a practical relaxation of this latent-variable model, where the binary variables $\mathbf{z}$ are replaced by their image-conditioned probabilities $\mathbf{u}$, yielding a continuous (soft) selection of radiomic features.

This view clarifies the relationships between the proposed method and the existing approaches summarised in Secs.~\ref{sec:preliminary_traditional_radiomics}~-~\ref{sec:end_to_end}:

\textit{\textbf{Classical radiomics without DL} }extract features and define ROI \emph{a priori}, and the classifier uses all (predefined) features. In our notation, this corresponds to $u_n\equiv 1$ for all $n$, with only $\boldsymbol{\beta}$ being learned from data (Eq.~\ref{eq_prob_radiomics}).

\textit{\textbf{Hybrid radiomics with DL}} may automate ROI delineation (Sec.~\ref{sec:preliminary_hybrid}) while keeping subsequent radiomic features and classical classification unchanged. This reduces manual effort but still yields a deterministic feature set per subject and does not explicitly model subject-specific feature usage probabilities.

\textit{\textbf{End-to-end DL approaches}} learn an implicit representation and prediction function $p(\mathbf{c}\mid\mathcal{I})=f_{\hat{\boldsymbol{\phi}}}(\mathcal{I})$ (Eq.~\ref{eq_prob_end2end}), without an explicit decomposition into radiomic evidence and feature selection, often making both feature- and case-level explanations less transparent.

Compared with these existing approaches, we discuss several advantages with the proposed individualised feature selection, as follows.

\textit{\textbf{Maintaining an explicit radiomic decision form:}}
The prediction is expressed directly in terms of interpretable radiomic features $\{r_n\}$ and a simple classifier with population-level parameters $\boldsymbol{\beta}$. This preserves the main appeal of classical radiomics, namely, that the decision can be inspected and attributed to known radiomic descriptors.

\textit{\textbf{Adding subject-specific, probabilistic feature usage:}}
Unlike classical and hybrid radiomics, the proposed model explicitly quantifies \emph{which} radiomic features are relied upon for each subject via $\mathbf{u}$. This provides an additional layer of interpretability: in addition to reporting effect sizes $\beta_n$, one can inspect the case-specific usage probabilities $u_n$ to understand the evidence actually used for an individual prediction.

\textit{\textbf{Improving expressibility without losing transparency:}}
The image-conditioned gating network $f_{\boldsymbol{\alpha}}$ enables the model to adapt feature usage to each subject, increasing flexibility beyond a single global linear radiomic model. At the same time, the final decision remains low-dimensional and explicit, avoiding the fully implicit representations of end-to-end DL.

\textit{\textbf{Automating of ROI/feature choice under weak supervision:}}
By predicting feature usage from the image using only condition labels, the model automates aspects of ROI and feature selection that are typically predefined in classical radiomics (Sec.~\ref{sec:preliminary_traditional_radiomics}), while retaining a transparent, radiomics-based explanation of the final decision.

\begin{figure}
    \centering
     \includegraphics[width=1\textwidth]{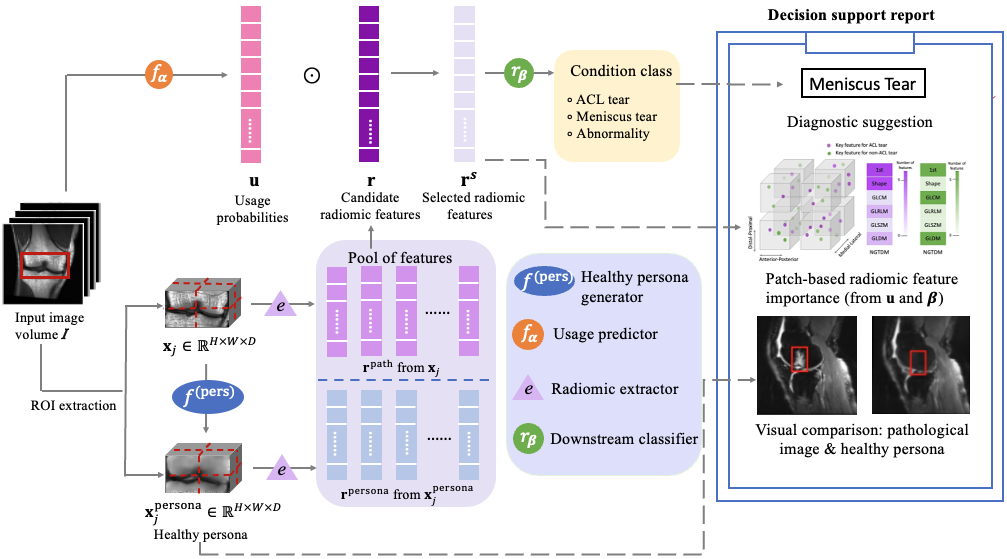}
    \caption{Overview of the radiomic feature selection framework (details in Sec.\ref{sec:method}).}
    \label{fig1}
\end{figure}

\subsection{Healthy persona and residual ROI-based radiomic features}
\label{sec:persona}

Pathological ROIs often contain both pathology-related cues and substantial nuisance variation (e.g., anatomy, scanner, demographics). We therefore introduce a healthy persona as an ROI-wise, subject-specific anatomical baseline, and represent pathology as deviations from this baseline, akin to residual learning~\cite{he2016deep} but on the ROI level.

For each ROI $\mathbf{x}_j$, we generate a corresponding healthy persona ROI:
\begin{equation}
\label{eq:persona_def}
\mathbf{x}_j^{\mathrm{persona}} = f^{(\mathrm{pers})}(\mathbf{x}_j)
\end{equation}
and define the residual ROI:
\begin{equation}
\label{eq:residual_def}
\Delta\mathbf{x}_j = \mathbf{x}_j - \mathbf{x}_j^{\mathrm{persona}},\quad j=1,\dots,J
\end{equation}
We implement $f^{(\mathrm{pers})}(\cdot)$ using a DDPM trained exclusively on healthy MRI images (details in Sec.~\ref{ddpm}).
Radiomic features are extracted per ROI using the same ROI-specific extractor $e^{(j)}(\cdot)$ as in Sec.~\ref{sec:preliminary_traditional_radiomics}:
\begin{equation}
\nonumber
\mathbf{r}^{\mathrm{path},(j)} = e^{(j)}(\mathbf{x}_j),\qquad
\mathbf{r}^{\mathrm{persona},(j)} = e^{(j)}(\mathbf{x}_j^{\mathrm{persona}}),\qquad
\mathbf{r}^{\Delta,(j)} = e^{(j)}(\Delta\mathbf{x}_j)
\end{equation}
These ROI-wise feature subvectors are then concatenated:
\begin{align}
\label{eq:persona_concat}
\mathbf{r}^{\mathrm{path}} &= \big[{\mathbf{r}^{\mathrm{path},(1)}}^\top,\dots,{\mathbf{r}^{\mathrm{path},(J)}}^\top\big]^\top\\
\mathbf{r}^{\mathrm{persona}} &= \big[{\mathbf{r}^{\mathrm{persona},(1)}}^\top,\dots,{\mathbf{r}^{\mathrm{persona},(J)}}^\top\big]^\top\\
\mathbf{r}^{\Delta} &= \big[{\mathbf{r}^{\Delta,(1)}}^\top,\dots,{\mathbf{r}^{\Delta,(J)}}^\top\big]^\top
\end{align}
Depending on the configuration, we use either
\begin{equation}
\label{eq:persona_feature_choice}
\mathbf{r}=\big[(\mathbf{r}^{\mathrm{path}})^\top, (\mathbf{r}^{\mathrm{persona}})^\top\big]^\top
 \quad\text{or}\quad
\mathbf{r}=\mathbf{r}^{\Delta}
\end{equation}
as the candidate radiomic feature vector in Sec.~\ref{sec:method_individualised}. This yields an explicit baseline (persona) and a deviation-focused representation (residual) while remaining compatible with the proposed individualised feature usage model.

\section{Experimental Results}

\subsection{Datasets and preprocessing}
\label{subsec:dataset}
\textit{\textbf{MRNet dataset:}}
We use the publicly available MRNet dataset~\cite{bien2018deep}, comprising 1,370 knee MRI examinations collected from axial, coronal and sagittal imaging planes. Each MRI scan in the dataset has been clinically annotated to indicate the presence or absence of abnormalities, specifically ACL tears and meniscal tears~\cite{bien2018deep}. In this study, we adopt the predefined dataset split of 1,130 cases for model training and 120 cases for validation to ensure comparability with existing literature and reproducibility of results.

\begin{table}[!ht]
\renewcommand{\thetable}{1}
\centering
\caption{Descriptive statistics of the MRNet dataset. Values in parentheses represent standard deviations or percentages, as appropriate.}
\label{tab4}

\renewcommand{\arraystretch}{1.5} 
\setlength{\tabcolsep}{6pt}
\fontsize{9pt}{10.5pt}\selectfont

\begin{tabular}{|p{5.5cm}|p{3cm}|}
\hline
\textbf{Statistic} & \textbf{Dataset} \\
\hline
\textbf{Number of exams} & 1130 \\
\hline
\textbf{Number of patients} & 1088 \\
\hline
\textbf{Age, mean (SD)} & 38.3 (16.9) \\
\hline
\textbf{Number of abnormality cases} (\%) & 913 (80.8) \\
\hline
\textbf{Number with ACL tear} (\%) & 208 (18.4) \\
\hline
\textbf{Number with meniscal tear} (\%) & 397 (35.1) \\
\hline
\textbf{Number with both ACL and meniscal tear} (\%) & 125 (11.1) \\
\hline
\end{tabular}
\end{table}

\textit{\textbf{ACL tear detection dataset:}}
We use the publicly available clinical dataset from the Clinical Hospital Centre Rijeka PACS~\cite{vstajduhar2017semi}, consisting of sagittal-plane knee MRI scans represented in 12-bit greyscale DICOM format. Each scan is accompanied by a clinically confirmed diagnosis concerning ACL tear status. 
In this study, we adopt the 664 cases for model training and 92 cases for validation. We perform a binary classification task distinguishing between ACL tear and non–ACL tear cases. This second dataset enables us to test model generalisability across institutions and acquisition protocols.

\textit{\textbf{Preprocessing:}} To facilitate uniformity and computational efficiency during model training and inference, all MRI scans undergo preprocessing involving resizing to standardized voxel dimensions of \( 32 \times 128 \times 128 \). 

To investigate the role of anatomical alignment on model training, we assess two preprocessing pipelines: (1) baseline spatial normalization without registration, and (2) affine spatial registration using NiftyReg~\cite{modat2010fast}, aligning each scan to a common reference space. This evaluation enables us to determine the impact of registration on feature consistency and persona reconstruction quality.

\subsection{Implementation details}

\subsubsection{Localised reconstruction model training}
\label{ddpm}

\begin{figure}
    \centering
     \includegraphics[width=0.9\textwidth]{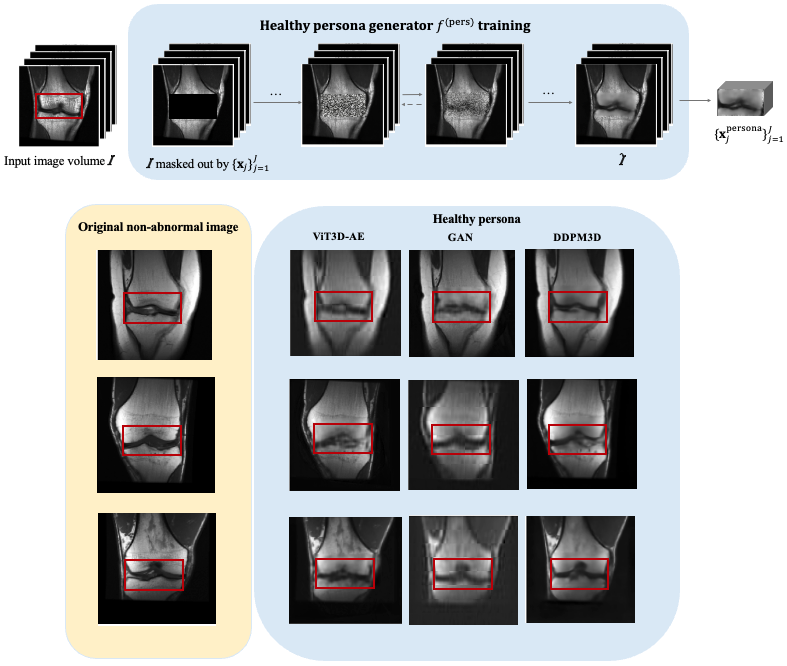}
    \caption{Top block: Workflow of DDPM3D-based reconstruction. Bottom block: comparison between original image and the generated healthy personas (details in Sec.\ref{sec:method}).} 
    \label{fig1}
\end{figure}

We implement a 3D DDPM for localised anatomical reconstruction, where the diffusion model operates on the full knee MRI volume $\mathcal{I}$ to generate a patient-specific healthy baseline. 
The DDPM is trained to inpaint the missing content and reconstruct a plausible pathology-free completion consistent with the observed
context, yielding a reconstructed healthy volume $\widehat{\mathcal{I}}$.

The forward diffusion process follows Ho et al.~\cite{ho2020denoising}, where Gaussian noise is incrementally added over \(T = 1000\) timesteps using a linear noise schedule \(\{\gamma_t\}_{t=1}^{T}\) ranging from \(1\times10^{-4}\) to \(2\times10^{-2}\).

The DDPM backbone is a 3D U-Net that operates on the full MRI volume $\mathcal{I}$, where the union of ROI
regions $\{\mathbf{x}_j\}_{j=1}^{J}$ is masked out to remove the corresponding voxel intensities. The model then
performs diffusion-based inpainting to reconstruct the healthy persona consistent with the surrounding
anatomy, producing a reconstructed healthy volume $\widehat{\mathcal{I}}$. We then obtain the healthy persona ROIs $\{\mathbf{x}^\mathrm{persona}_j\}_{j=1}^{J}$ for each region by cropping the reconstructed volume to the same ROI location. 
The training objective minimises the mean squared error (MSE) between the reconstructed and ground-truth healthy volumes (equivalently, over the masked region), encouraging accurate anatomical inpainting:
\(
\mathcal{L}_{\text{recon}} = \left\| \mathbf{x}_j^{\mathrm{persona}} - \mathbf{x}_j \right\|^2
\)

The model is trained on healthy MRI scans from the MRNet dataset (cases labelled \(abnormal = 0\)), consisting of 654 volumes from 218 patients across three views. Each scan was masked by a central bounding box, with a \(50\%\) depth, a \(30\%\) height and a \(50\%\) width of the original image size. The remaining unmasked context provides spatial cues for the reconstruction task.

We compare the reconstruction performance of DDPM3D against two baseline models: a 3D GAN and a ViT3D-based autoencoder (ViT3D-AE). All models are trained under the same configuration, using identical ROI sampling region (\(50\%\) depth, a \(30\%\) height and a \(50\%\) width of the original image size), view settings, and preprocessing with registration. As shown in Table~\ref{tab2}, DDPM3D achieves the best reconstruction fidelity, with a mean SSIM of 0.92 and an MSE of 104.53, significantly outperforming both ViT3D-AE and GAN (all p-values $<0.05$). The superior performance of DDPM3D suggests that denoising diffusion models are more effective at modelling complex anatomical structures from limited contextual input, producing reconstructions that are both sharper and more anatomically consistent (Fig.~\ref{ddpm}). In contrast, the GAN-based approach, while capable of generating plausible textures, exhibits greater variability and higher reconstruction error. The ViT3D-AE, despite capturing global features well, appears to underperform in fine-grained anatomical recovery compared to the diffusion-based model. These results highlight the advantages of probabilistic, iterative refinement in DDPM3D over the deterministic or adversarial learning paradigms employed by the baselines.

\begin{table}[!ht]
\renewcommand{\thetable}{2}
\centering
\caption{Comparison study for reconstruction model. Reg: Registration in preprocessing. 
}
\label{tab2}

\renewcommand{\arraystretch}{1.8} 
\setlength{\tabcolsep}{4pt}
\fontsize{9pt}{10pt}\selectfont

\begin{tabular}{|c|c|c|c|cc|}
\hline
\textbf{Method} & \textbf{ROI} & \textbf{View} & \textbf{Reg} & \multicolumn{2}{c|}{\textbf{Evaluation Metrics}} \\
\cline{5-6}
& & & & \textbf{MSE} & \textbf{SSIM} \\
\hline

\textbf{DDPM3D} & 0.5×0.3×0.5 & 3 & \ding{51} & \textbf{104.53$\pm$28.19} & \textbf{0.92$\pm$0.01} \\
\hline

\textbf{ViT3D-AE} & 0.5×0.3×0.5 & 3 & \ding{51} & 168.19$\pm$7.84 & 0.87$\pm$0.01 \\
\hline

\textbf{GAN} & 0.5×0.3×0.5 & 3 & \ding{51} & 144.87$\pm$33.24 & 0.84$\pm$0.02 \\
\hline

\end{tabular}
\end{table}

By training exclusively on healthy MRI scans, DDPM3D is expected to effectively inpaint pathological ROIs with anatomically realistic structures, facilitating robust reconstruction for downstream radiomic analysis. We do not aim to recover the true pathological tissue; rather, the inpainted persona provides a consistent healthy baseline, which we show experimentally is useful enough for robust downstream radiomic analysis.

\begin{table}[!ht]
\renewcommand{\thetable}{3}
\centering
\caption{Ablation study of DDPM3D model configurations for 3D reconstruction. Reg: Registration in preprocessing.}
\label{tab2}

\renewcommand{\arraystretch}{1.8} 
\setlength{\tabcolsep}{1.8pt}
\fontsize{9pt}{10pt}\selectfont

\begin{tabular}{|c|c|c|c|c|c|c|}
\hline
\textbf{Method} & \textbf{Location} & \textbf{ROI} & \textbf{View} & \textbf{Reg} & \multicolumn{2}{c|}{\textbf{Evaluation Metrics}} \\
\cline{6-7}
 & & & & & \textbf{SSIM} & \textbf{MSE} \\
\hline

\multirow{10}{*}{\textbf{DDPM3D}} 
& Central & 0.5×0.3×0.5 & 3              & \ding{51} & \textbf{0.92$\pm$0.01} & \textbf{111.53$\pm$28.19} \\
& Central & 0.5×0.3×0.5 & 1 (sagittal)   & \ding{51} & 0.86$\pm$0.10 & 197.07$\pm$106.50 \\
& Central & 0.5×0.3×0.5 & 1 (coronal)    & \ding{51} & 0.92$\pm$0.07 & 135.52$\pm$33.14 \\
& Central & 0.5×0.3×0.5 & 1 (axial)      & \ding{51} & 0.92$\pm$0.10 & 192.52$\pm$112.76 \\
& Central & 0.3×0.3×0.3 & 3              & \ding{51} & 0.92$\pm$0.10 & 111.53$\pm$18.19 \\
& Central & 0.6×0.6×0.6 & 3              & \ding{51} & 0.61$\pm$0.07 & 752.24$\pm$245.66 \\
& Central & 0.8×0.8×0.8 & 3              & \ding{51} & 0.61$\pm$0.07 & 812.03$\pm$205.72 \\
& Random  & 0.5×0.3×0.5 & 3              & \ding{55} & 0.90$\pm$0.10 & 155.20$\pm$128.57 \\
& Random  & 0.5×0.3×0.5 & 1 (coronal)    & \ding{55} & 0.90$\pm$0.08 & 229.02$\pm$141.95 \\
\hline
\end{tabular}
\end{table}

\subsubsection{Feature extraction}
\label{sec:Feature Extraction}
We extract radiomic features from both pathological MRI scans and their corresponding DDPM-generated healthy personas. Feature extraction is performed using a PyTorch-adapted implementation of the PyRadiomics library~\cite{van2017computational}. The complete list of implemented radiomic features and their definitions are available in the official PyRadiomics documentation.

Specifically, we include a set of radiomic feature families:  
(1) \textit{First-Order Statistics}, describing voxel intensity distributions (e.g., mean, skewness, entropy);  
(2) \textit{3D Shape Features}, capturing volumetric and geometric descriptors (e.g., elongation, surface area, compactness).

Both the pathological image and its healthy persona are divided into a regular grid of \(n \times n \times n\) subpatches, where \(n\) controls the level of spatial granularity. We systematically evaluate three decomposition settings: \(1 \times 1 \times 1\), \(2 \times 2 \times 2\) and \(3 \times 3 \times 3\), corresponding to 1, 8 and 27 subpatches per volume, respectively. Candidate radiomic features are extracted independently from each ROI patch across three anatomical planes, and from both image types (pathological and healthy persona). In addition to the 35 candidate radiomic features, we include 3 discrete spatial indices for each subpatch, reflecting its relative location within the \(n^3\) grid.

Thus, each subpatch contributes a total of \(35 + 3 = 38\) candidate radiomic features. For a given configuration \(n\), the patient-level feature dimensionality of  
\[
\dim(\mathbf{r}) \;=\;3 \times (n^3) \times 2 \times (38)
\]

\noindent where \(3\) is the number of anatomical views, \(n^3\) is the number of subpatches, \(2\) corresponds to the two image types (pathological and reconstructed), and 38 is the number of candidate radiomic features per subpatch.

\subsubsection{Feature selection}
We employ a 3DResNet-18~\cite{he2016deep} as the usage predictor \(f_{\boldsymbol{\alpha}}(\cdot)\) to estimate subject-specific feature usage probabilities (i.e., usage weights) \(\mathbf{u}\in[0,1]^N\) over the candidate radiomic features. The network takes the full MRI volume (three anatomical views) as input and uses a pretrained ResNet-18 backbone (classification head removed) followed by dropout, a fully connected layer with layer normalisation, and a sigmoid output unit to produce \(\mathbf{u}\). The usage predictor is trained jointly with the downstream classifier using binary cross-entropy with positive-class weighting to address class imbalance, optimised with Adam.

\subsubsection{Classification model:}
\label{classifier}
During training, the usage-weighted candidate radiomic feature is fed into the logistic regression classifier to predict class probabilities for the downstream classification task. The classifier is optimised using binary cross-entropy loss with positive class weighting. 
After training, we select the decision threshold on the validation set by maximising Youden’s \(J\) statistic, \(J=\mathrm{Sensitivity}+\mathrm{Specificity}-1.\) In our implementation, the threshold is re-estimated every 10 epochs from the current validation predictions. During inference, hard feature selection is performed by thresholding the usage weights, with threshold values empirically selected for each application to balance classification performance and feature sparsity (the threshold values are reported in Tables~\ref{tab1}–\ref{tab3})).

\subsection{Comparison studies}

Comprehensive comparisons were conducted against established end-to-end DL baselines, including MRNet~\cite{bien2018deep} and ELNet~\cite{tsai2020knee}. These models are widely used in musculoskeletal MRI analysis and serve as strong references for evaluating both classification performance and interpretability. Reported metrics include Accuracy (Acc), Sensitivity (Sen), Specificity (Spe) and Area Under the Curve (AUC), evaluated across multiple disease categories: abnormality (abn), anterior cruciate ligament tear (acl) and meniscus tear (men).

\vspace{1ex}
\textit{\noindent\textbf{Performance on MRNet dataset.} }Table~\ref{tab1} presents results on the MRNet dataset. Using a finer grid ($3\times3\times3$) yields the highest AUC for ACL and meniscus tear detection, consistent with these tasks being driven by focal, spatially localised imaging cues that benefit from higher spatial granularity. In contrast, the $2 \times 2 \times 2$ setup performed better for abnormality detection, suggesting that coarser representations may be more robust for global anomalies. 

Our proposed method with $2 \times 2 \times 2$ subpatches is significantly better than ELNet and MRNet in meniscus tear detection (\textit{p}-values $<0.050$ for accuracy, sensitivity and specificity). Our proposed method with $3 \times 3 \times 3$ subpatches surpasses the ELNet and MRNet  in meniscus tear detection all evaluated metrics (\textit{p}-values $<0.050$). Our method exhibits high sensitivity while maintaining comparable specificity, demonstrating comparable or even superior performance across multiple configurations. 

\begin{table}[!ht]
\renewcommand{\thetable}{5}

\centering
\caption{Comparison study on MRNet dataset. N: Number of subpatches; UP: Radiomic feature usage predictor; T: Application-specific threshold hyperparameter; Reg: Registration in preprocessing. 
* denotes reproduced results}
\label{tab1}

\renewcommand{\arraystretch}{1.5} 

\setlength{\tabcolsep}{1.3pt}
\fontsize{9pt}{10pt}\selectfont

\begin{tabular}{|@{\hskip 1pt}c|@{\hskip 1pt}c|@{\hskip 1pt}c|@{\hskip 1pt}c|@{\hskip 1pt}c|@{\hskip 1pt}c|@{\extracolsep{1.5pt}}c|c@{\hskip 1.5pt}c@{\hskip 1.5pt}c@{\hskip 1.5pt}c|}

\hline
 & & & & & & & \multicolumn{4}{c|}{\textbf{Evaluation Metrics}} \\ 
\cline{8-11} 
\textbf{Method} & \textbf{N} & \textbf{UP} & \textbf{Persona} & \textbf{T} & \textbf{Reg} & \textbf{Type} & \textbf{Acc} & \textbf{Sen} & \textbf{Spe} & \textbf{AUC} \\ 
\hline

\hline
\textbf{\multirow{3}{*}{Ours}} &\multirow{3}{*}{2*2*2} & \multirow{3}{*}{\ding{51}} & \multirow{3}{*}{\ding{51}} & \multirow{3}{*}{0.4} & \multirow{3}{*}{\ding{51}} 
& abn & \textbf{0.90$\pm$0.13} & 0.94$\pm$0.10 & 0.77$\pm$0.25 & 0.85$\pm$0.16 \\
&&&&&& acl & 0.81$\pm$0.10 & 0.92$\pm$0.08 & 0.66$\pm$0.23 & 0.80$\pm$0.12 \\
&&&&&& men & 0.82$\pm$0.11 & 0.81$\pm$0.14 & 0.88$\pm$0.14 & 0.84$\pm$0.11 \\
\hline

\textbf{\multirow{3}{*}{Ours}} & \multirow{3}{*}{3*3*3} & \multirow{3}{*}{\ding{51}} & \multirow{3}{*}{\ding{51}} & \multirow{3}{*}{0.4} & \multirow{3}{*}{\ding{51}} 
& abn & 0.84$\pm$0.12 & 0.96$\pm$0.08 & 0.54$\pm$0.33 & 0.75$\pm$0.25 \\
&&&&&& acl & \textbf{0.87$\pm$0.16} & 0.88$\pm$0.13 & 0.87$\pm$0.22 & 0.87$\pm$0.15 \\
&&&&&& men & \textbf{0.87$\pm$0.23} & 0.88$\pm$0.16 & 0.87$\pm$0.30 & 0.88$\pm$0.20 \\
\hline

\multirow{3}{*}{\textbf{ELNet*}} & \multirow{3}{*}{-} & \multirow{3}{*}{-} & \multirow{3}{*}{-} & \multirow{3}{*}{-} & \multirow{3}{*}{-} 
& abn & 0.80$\pm$0.01 & 0.95$\pm$0.01 & 0.22$\pm$0.03 & 0.73$\pm$0.01 \\
&&&&&& acl & 0.70$\pm$0.10 & 0.72$\pm$0.26 & 0.68$\pm$0.04 & 0.73$\pm$0.08 \\
&&&&&& men & 0.65$\pm$0.06 & 0.61$\pm$0.13 & 0.63$\pm$0.07 & 0.69$\pm$0.03 \\
\hline
\multirow{3}{*}{\textbf{MRNet*}} & \multirow{3}{*}{-} & \multirow{3}{*}{-} & \multirow{3}{*}{-} & \multirow{3}{*}{-} & \multirow{3}{*}{-} 
& abn & 0.85$\pm$0.01 & 0.84$\pm$0.03 & 0.65$\pm$0.09 & 0.94$\pm$0.02 \\
&&&&&& acl & 0.86$\pm$0.02 & 0.77$\pm$0.02 & 0.97$\pm$0.02 & 0.97$\pm$0.02 \\
&&&&&& men & 0.71$\pm$0.04 & 0.69$\pm$0.03 & 0.74$\pm$0.03 & 0.84$\pm$0.04 \\

\hline


\end{tabular}
\end{table}

\vspace{1ex}
\textit{\noindent\textbf{Performance on ACL tear detection dataset.}} To further validate the framework, an additional ACL-specific dataset was analyzed (Table~\ref{tab2}). Again, the $3 \times 3 \times 3$ configuration outperformed baselines with an AUC of 0.85 and accuracy of 0.88, confirming the value of patient-specific radiomic features usage prediction and personalised healthy persona. Compared to MRNet and ELNet, the proposed method with $3 \times 3 \times 3$ configuration achieves better performance (\textit{p}-values $<0.050$ for accuracy and sensitivity), which are critical for reducing false negatives in clinical settings.

\begin{table}[!ht]
\renewcommand{\thetable}{6}
\centering
\caption{Comparison study on ACL tear detection dataset. N: Number of subpatches; UP: Radiomic feature usage predictor; T: Application-specific threshold hyperparameter; Reg: Registration in preprocessing. 
* denotes reproduced results}
\label{tab2}

\renewcommand{\arraystretch}{1.8} 
\setlength{\tabcolsep}{1.8pt}
\fontsize{9pt}{10pt}\selectfont

\begin{tabular}{|@{\hskip 1.5pt}c|@{\hskip 1.5pt}c|@{\hskip 1.5pt}c|@{\hskip 1.5pt}c|@{\hskip 1.5pt}c|@{\hskip 1.5pt}c|@{\hskip 1.5pt}c@{\extracolsep{1.5pt}}c@{\hskip 1.5pt}c@{\hskip 1.5pt}c|}
\hline
 & & & & & & \multicolumn{4}{c|}{\textbf{Evaluation Metrics}} \\ 
\cline{7-10} 
\textbf{Method} & \textbf{N} & \textbf{UP} & \textbf{Persona} & \textbf{T} & \textbf{Reg} & \textbf{Acc} & \textbf{Sen} & \textbf{Spe} & \textbf{AUC} \\ 
\hline

\textbf{Ours} & 2*2*2 & \ding{51} & \ding{51} & 0.5 & \ding{51} 
& 0.87$\pm$0.06 & 0.87$\pm$0.12 & 0.59$\pm$0.14 & 0.81$\pm$0.09 \\
\hline

\textbf{Ours} & 3*3*3 & \ding{51} & \ding{51} & 0.5 & \ding{51} 
& \textbf{0.88$\pm$0.02} & 0.95$\pm$0.10 & 0.58$\pm$0.10 & \textbf{0.85$\pm$0.08} \\
\hline

\textbf{ELNet*} & - & - & - & - & - 
& 0.84$\pm$0.00 & 0.73$\pm$0.03 & 0.61$\pm$0.02 & 0.83$\pm$0.01 \\
\hline

\textbf{MRNet*} & - & - & - & - & - 
& 0.78$\pm$0.06 & 0.52$\pm$0.02 & 0.58$\pm$0.19 & 0.80$\pm$0.12 \\
\hline

\end{tabular}
\end{table}


\subsection{Ablation study}
To evaluate the contribution of key components, we conducted controlled ablation experiments on the MRNet dataset (Table~\ref{tab3}). The $2 \times2\times2$ setting yielded the best balance between performance and interpretability. Finer candidate ROIs  ($3\times3\times3$) improved AUC for localised tasks (e.g., ACL, meniscus) but introduced variability, while coarser larger candidate ROIs ($1\times1\times1$) underperformed across all metrics. Removing the usage predictor network led to reduced specificity and overall AUC, confirming the benefit of adaptive, subject-specific feature usage probabilities. Excluding the healthy persona reduced performance (when the feature pool include the first order and shape features), most noticeably for ACL. Omitting image registration significantly degraded sensitivity. These results validate each component’s contribution to the model’s accuracy and interpretability.

\begin{table}[!ht]

\renewcommand{\thetable}{7}

\centering
\caption{Comparison of different configurations on MRNet dataset. N: Number of subpatches; UP: Radiomic feature usage predictor; T: Application-specific threshold hyperparameter; Reg: Registration in preprocessing. 
* denotes reproduced results}
\label{tab3}

\renewcommand{\arraystretch}{1.5} 

\setlength{\tabcolsep}{1.3pt}
\fontsize{9pt}{10pt}\selectfont

\begin{tabular}{|@{\hskip 1pt}c|@{\hskip 1pt}c|@{\hskip 1pt}c|@{\hskip 1pt}c|@{\hskip 1pt}c|@{\hskip 1pt}c|@{\extracolsep{1.5pt}}c|c@{\hskip 1.5pt}c@{\hskip 1.5pt}c@{\hskip 1.5pt}c|}

\hline
 & & & & & & & \multicolumn{4}{c|}{\textbf{Evaluation Metrics}} \\ 
\cline{8-11} 
\textbf{Method} & \textbf{N} & \textbf{UP} & \textbf{Persona} & \textbf{T} & \textbf{Reg} & \textbf{Type} & \textbf{Acc} & \textbf{Sen} & \textbf{Spe} & \textbf{AUC} \\ 
\hline

\hline
\textbf{\multirow{3}{*}{Ours}} &\multirow{3}{*}{2*2*2} & \multirow{3}{*}{\ding{51}} & \multirow{3}{*}{\ding{51}} & \multirow{3}{*}{0.4} & \multirow{3}{*}{\ding{51}} 
& abn & \textbf{0.90$\pm$0.13} & 0.94$\pm$0.10 & 0.77$\pm$0.25 & 0.85$\pm$0.16 \\
&&&&&& acl & 0.81$\pm$0.10 & 0.92$\pm$0.08 & 0.66$\pm$0.23 & 0.80$\pm$0.12 \\
&&&&&& men & 0.82$\pm$0.11 & 0.81$\pm$0.14 & 0.88$\pm$0.14 & 0.84$\pm$0.11 \\
\hline

\textbf{\multirow{3}{*}{Ours}} & \multirow{3}{*}{3*3*3} & \multirow{3}{*}{\ding{51}} & \multirow{3}{*}{\ding{51}} & \multirow{3}{*}{0.4} & \multirow{3}{*}{\ding{51}} 
& abn & 0.84$\pm$0.12 & 0.96$\pm$0.08 & 0.54$\pm$0.33 & 0.75$\pm$0.25 \\
&&&&&& acl & \textbf{0.87$\pm$0.16} & 0.88$\pm$0.13 & 0.87$\pm$0.22 & 0.87$\pm$0.15 \\
&&&&&& men & \textbf{0.87$\pm$0.23} & 0.88$\pm$0.16 & 0.87$\pm$0.30 & 0.88$\pm$0.20 \\
\hline
\multirow{3}{*}{T=0.0} & \multirow{3}{*}{2*2*2} & \multirow{3}{*}{\ding{51}} & \multirow{3}{*}{\ding{51}} & \multirow{3}{*}{0.0} & \multirow{3}{*}{\ding{51}} 
& abn & \textbf{0.90$\pm$0.11} & 0.99$\pm$0.03 & 0.63$\pm$0.42 & 0.81$\pm$0.21 \\
&&&&&& acl & 0.84$\pm$0.13 & 0.83$\pm$0.23 & 0.84$\pm$0.16 & 0.83$\pm$0.15 \\
&&&&&& men & \textbf{0.83$\pm$0.16} & 0.82$\pm$0.26 & 0.83$\pm$0.13 & 0.83$\pm$0.17 \\
\hline
\multirow{3}{*}{T=0.5} & \multirow{3}{*}{2*2*2} & \multirow{3}{*}{\ding{51}} & \multirow{3}{*}{\ding{51}} & \multirow{3}{*}{0.5} & \multirow{3}{*}{\ding{51}} 
& abn & 0.76$\pm$0.18 & 0.72$\pm$0.22 & 0.90$\pm$0.15 & 0.81$\pm$0.17 \\
&&&&&& acl & 0.75$\pm$0.17 & 0.79$\pm$0.18 & 0.71$\pm$0.36 & 0.75$\pm$0.15 \\
&&&&&& men & 0.78$\pm$0.13 & 0.74$\pm$0.11 & 0.85$\pm$0.17 & 0.82$\pm$0.09 \\
\hline
\multirow{3}{*}{1patch}& \multirow{3}{*}{1*1*1} & \multirow{3}{*}{\ding{51}} & \multirow{3}{*}{\ding{51}} & \multirow{3}{*}{0.4} & \multirow{3}{*}{\ding{51}} 
& abn & 0.70$\pm$0.35 & 0.64$\pm$0.46 & 0.90$\pm$0.15 & 0.78$\pm$0.23 \\
&&&&&& acl & 0.73$\pm$0.15 & 0.96$\pm$0.06 & 0.56$\pm$0.26 & 0.76$\pm$0.13 \\
&&&&&& men & 0.60$\pm$0.08 & 0.76$\pm$0.27 & 0.50$\pm$0.28 & 0.64$\pm$0.05 \\
\hline
\multirow{3}{*}{NoPersona}& \multirow{3}{*}{2*2*2} & \multirow{3}{*}{\ding{51}} & \multirow{3}{*}{\ding{55}} & \multirow{3}{*}{0.4} & \multirow{3}{*}{\ding{51}} 
& abn & 0.77$\pm$0.13 & 0.75$\pm$0.19 & 0.87$\pm$0.18 & 0.81$\pm$0.10 \\
&&&&&& acl & 0.66$\pm$0.18 & 0.92$\pm$0.10 & 0.45$\pm$0.36 & 0.68$\pm$0.15 \\
&&&&&& men & 0.72$\pm$0.10 & 0.80$\pm$0.15 & 0.68$\pm$0.10 & 0.74$\pm$0.10 \\
\hline
\multirow{3}{*}{NoFS}& \multirow{3}{*}{2*2*2} & \multirow{3}{*}{\ding{55}} & \multirow{3}{*}{\ding{51}} & \multirow{3}{*}{0.4} & \multirow{3}{*}{\ding{51}} 
& abn & 0.81$\pm$0.03 & 0.93$\pm$0.05 & 0.53$\pm$0.22 & 0.68$\pm$0.09 \\
&&&&&& acl & 0.77$\pm$0.11 & 0.83$\pm$0.12 & 0.60$\pm$0.32 & 0.72$\pm$0.14 \\
&&&&&& men & 0.71$\pm$0.05 & 0.78$\pm$0.07 & 0.66$\pm$0.11 & 0.72$\pm$0.03 \\
\hline
\multirow{3}{*}{NoReg}& \multirow{3}{*}{2*2*2} & \multirow{3}{*}{\ding{51}} & \multirow{3}{*}{\ding{51}} & \multirow{3}{*}{0.4} & \multirow{3}{*}{\ding{55}} 
& abn & 0.77$\pm$0.08 & 0.57$\pm$0.13 & 0.96$\pm$0.04 & 0.76$\pm$0.06 \\
&&&&&& acl & 0.73$\pm$0.10 & 0.40$\pm$0.29 & 0.99$\pm$0.03 & 0.70$\pm$0.10 \\
&&&&&& men & 0.67$\pm$0.15 & 0.99$\pm$0.03 & 0.41$\pm$0.27 & 0.70$\pm$0.14 \\


\hline

\end{tabular}
\end{table}

\textit{\noindent\textbf{Impact of feature families and healthy persona.} } To further investigate the role of different candidate radiomic feature families and the contribution of the healthy persona, we extended the candidate radiomic features to include higher-order texture features from PyRadiomics, namely the grey Level Co-occurrence Matrix (GLCM), grey Level Run Length Matrix (GLRLM), grey Level Size Zone Matrix (GLSZM), Neighbouring grey Tone Difference Matrix (NGTDM) and grey Level Dependence Matrix (GLDM). In this setting, each subpatch contributed 113 candidate radiomic features (including spatial location indices), resulting in a subject-level feature dimensionality of
\[
\dim(\mathbf{r}) \;=\; 3 \times (n^3) \times 2 \times (113).
\]

We evaluated three configurations: (1) the proposed \emph{First-Order + Shape} feature set, (2) \emph{All Radiomic Features} (First-Order, Shape and Higher-Order Texture), and (3) \emph{All Radiomic Features + Healthy Persona}. The results are presented in Table~\ref{tab1}.  

Compared with the proposed First-Order + Shape baseline, the inclusion of higher-order texture candidate radiomic features increased sensitivity for abnormality and meniscus (abn: $0.94\pm0.10 \rightarrow 0.98\pm0.03$; men: $0.81\pm0.14 \rightarrow 0.97\pm0.04$) and improved AUC for ACL (ACL: $0.80\pm0.12 \rightarrow 0.92\pm0.04$) . Although these findings suggest that texture-based radiomics can capture complementary information on intra-tissue heterogeneity and spatial grey-level dependencies, such features are generally less interpretable than First-Order and Shape featuress~\cite{zhang2023artificial}, which may limit their utility in clinical decision support.  

When the healthy persona was incorporated alongside the full radiomic feature set, it did not consistently improve performance over the full radiomic feature set alone. For abnormality and ACL, \emph{All RF + Persona} reduced specificity and AUC relative to \emph{All RF-NoPersona} (abn: Spe $0.71\pm0.33 \rightarrow 0.54\pm0.41$, AUC $0.85\pm0.17 \rightarrow 0.76\pm0.21$; ACL: Spe $0.88\pm0.05 \rightarrow 0.70\pm0.21$, AUC $0.92\pm0.04 \rightarrow 0.84\pm0.10$), while meniscus sensitivity increased ($0.97\pm0.04 \rightarrow 1.00\pm0.00$).This behavior indicates that the healthy persona emphasizes pathological deviations and improved interpretability, but at the cost of potential over-sensitivity to subtle image variations.  

These results confirm that the proposed configuration offers a strong and interpretable foundation, while extended variants demonstrate the trade-off between improved sensitivity and reduced interpretability. The choice between configurations may therefore depend on the relative priority of interpretability versus maximal detection sensitivity in a given clinical application.

\begin{table}[!ht]

\renewcommand{\thetable}{8}

\centering
\caption{Comparison study evaluating impact of higher order radiomic features group. N: Number of patches; RF: Radiomic features; UP: Radiomic feature usage predictor; Reg: Registration in preprocessing. 
}
\label{higher}

\renewcommand{\arraystretch}{1.5} 

\setlength{\tabcolsep}{1.3pt}
\fontsize{8pt}{9pt}\selectfont

\begin{tabular}{|@{\hskip 0.9pt}c|@{\hskip 0.9pt}c|@{\hskip 0.9pt}c|@{\hskip 0.9pt}c|@{\hskip 0.9pt}c|@{\hskip 0.9pt}c|@{\extracolsep{1.5pt}}c|c@{\hskip 1.5pt}c@{\hskip 1.5pt}c@{\hskip 1.5pt}c|}

\hline
 & & & & & & & \multicolumn{4}{c|}{\textbf{Evaluation Metrics}} \\ 
\cline{8-11} 
\textbf{Method} & \textbf{N} & \textbf{2nd RF} & \textbf{UP} & \textbf{Persona} & \textbf{Reg} & \textbf{Type} & \textbf{Acc} & \textbf{Sen} & \textbf{Spe} & \textbf{AUC} \\ 
\hline

\hline
\textbf{\multirow{3}{*}{Ours}} &\multirow{3}{*}{2*2*2} & \multirow{3}{*}{\ding{55}} & \multirow{3}{*}{\ding{51}} & \multirow{3}{*}{\ding{51}} & \multirow{3}{*}{\ding{51}} 
& abn & \textbf{0.90$\pm$0.13} & 0.94$\pm$0.10 & 0.77$\pm$0.25 & 0.85$\pm$0.16 \\
&&&&&& acl & 0.81$\pm$0.10 & 0.92$\pm$0.08 & 0.66$\pm$0.23 & 0.80$\pm$0.12 \\
&&&&&& men & 0.82$\pm$0.11 & 0.81$\pm$0.14 & 0.88$\pm$0.14 & 0.84$\pm$0.11 \\
\hline
\textbf{\multirow{3}{*}{Ours}} & \multirow{3}{*}{3*3*3} & \multirow{3}{*}{\ding{55}} & \multirow{3}{*}{\ding{51}} & \multirow{3}{*}{\ding{51}} & \multirow{3}{*}{\ding{51}} 
& abn & 0.84$\pm$0.12 & 0.96$\pm$0.08 & 0.54$\pm$0.33 & 0.75$\pm$0.25 \\
&&&&&& acl & \textbf{0.87$\pm$0.16} & 0.88$\pm$0.13 & 0.87$\pm$0.22 & 0.87$\pm$0.15 \\
&&&&&& men & \textbf{0.87$\pm$0.23} & 0.88$\pm$0.16 & 0.87$\pm$0.30 & 0.88$\pm$0.20 \\
\hline

\textbf{\multirow{3}{*}{all RF}} & \multirow{3}{*}{2*2*2} & \multirow{3}{*}{\ding{51}} & \multirow{3}{*}{\ding{51}} & \multirow{3}{*}{\ding{51}} & \multirow{3}{*}{\ding{51}} 
& abn & 0.87$\pm$0.11 & 0.98$\pm$0.03 & 0.54$\pm$0.41 & 0.76$\pm$0.21 \\
&&&&&& acl & 0.83$\pm$0.11 & 0.98$\pm$0.04 & 0.70$\pm$0.21 & 0.84$\pm$0.10 \\
&&&&&& men & 0.83$\pm$0.24 & 1.00$\pm$0.00 & 0.73$\pm$0.46 & 0.87$\pm$0.23 \\

\hline
\textbf{\multirow{3}{*}{all RF-NoPersona}} &\multirow{3}{*}{2*2*2} & \multirow{3}{*}{\ding{51}} & \multirow{3}{*}{\ding{51}} & \multirow{3}{*}{\ding{55}} & \multirow{3}{*}{\ding{51}} 
& abn & \textbf{0.92$\pm$0.09} & 0.98$\pm$0.03 & 0.71$\pm$0.33 & 0.85$\pm$0.17 \\
&&&&&& acl & \textbf{0.94$\pm$0.09} & 0.96$\pm$0.06 & 0.88$\pm$0.05 & 0.92$\pm$0.04 \\
&&&&&& men & 0.84$\pm$0.16 & 0.97$\pm$0.04 & 0.67$\pm$0.34 & 0.82$\pm$0.18 \\
\hline

\end{tabular}
\end{table}

\section{Interpretability and Clinical Insights}

\subsection{Summary of ACL tears radiomic feature analysis} 
For ACL injuries, key features are concentrated in the sagittal view and the medial/posterior regions of the ROI. This aligns with both clinical expectations, as the sagittal view provides the clearest visualization of ACL continuity, and anatomically, the ACL originates from the intercondylar eminence of the tibia and inserts onto the medial aspect of the lateral femoral condyle\cite{petersen2007anatomy}. The key radiomic features in the medial and posterior subpatches for ACL tears include \emph{Entropy}, \emph{Interquartile Range} and \emph{Compactness}. Entropy quantifies the complexity and irregularity in the image intensity patterns, typically increasing when torn ligament fibers appear irregular and heterogeneous. Interquartile Range, representing the spread of voxel intensities in the middle half of the patch, captures the intensity heterogeneity indicative of discontinuous fibers. Compactness, which quantifies how closely a shape approximates a sphere, reflects structural deformation and may highlight secondary signs such as posterior cruciate ligament laxity and anterior tibial translation. 


\subsection{Summary of meniscus tears radiomic feature analysis} 
Key features for meniscus tear are not localised to specific regions, as lesions can occur in either or both menisci. On PD-weighted MRI, these tears characteristically appear as a linear high-intensity signal coursing through the meniscus, typically extending to its superior or inferior surface. Key radiomic features in the lateral subpatches include \emph{Energy}, which measures overall tissue integrity and decreases with disruption of the meniscal substance, \emph{Uniformity}, which detects irregular voxel intensities in torn regions, and the \emph{10percentile} feature, 
which highlights subtle changes in the lower-intensity range of the meniscal signal. Given the variability in tear location and morphology, patient-specific radiomic feature selection remains essential for robust detection and quantification of meniscal pathology.

\subsection{Case 1: medial bucket-handle meniscal tear with partial ACL rupture}
\label{case1paragraph}
\begin{figure}
    \centering

     \includegraphics[width=1\textwidth]{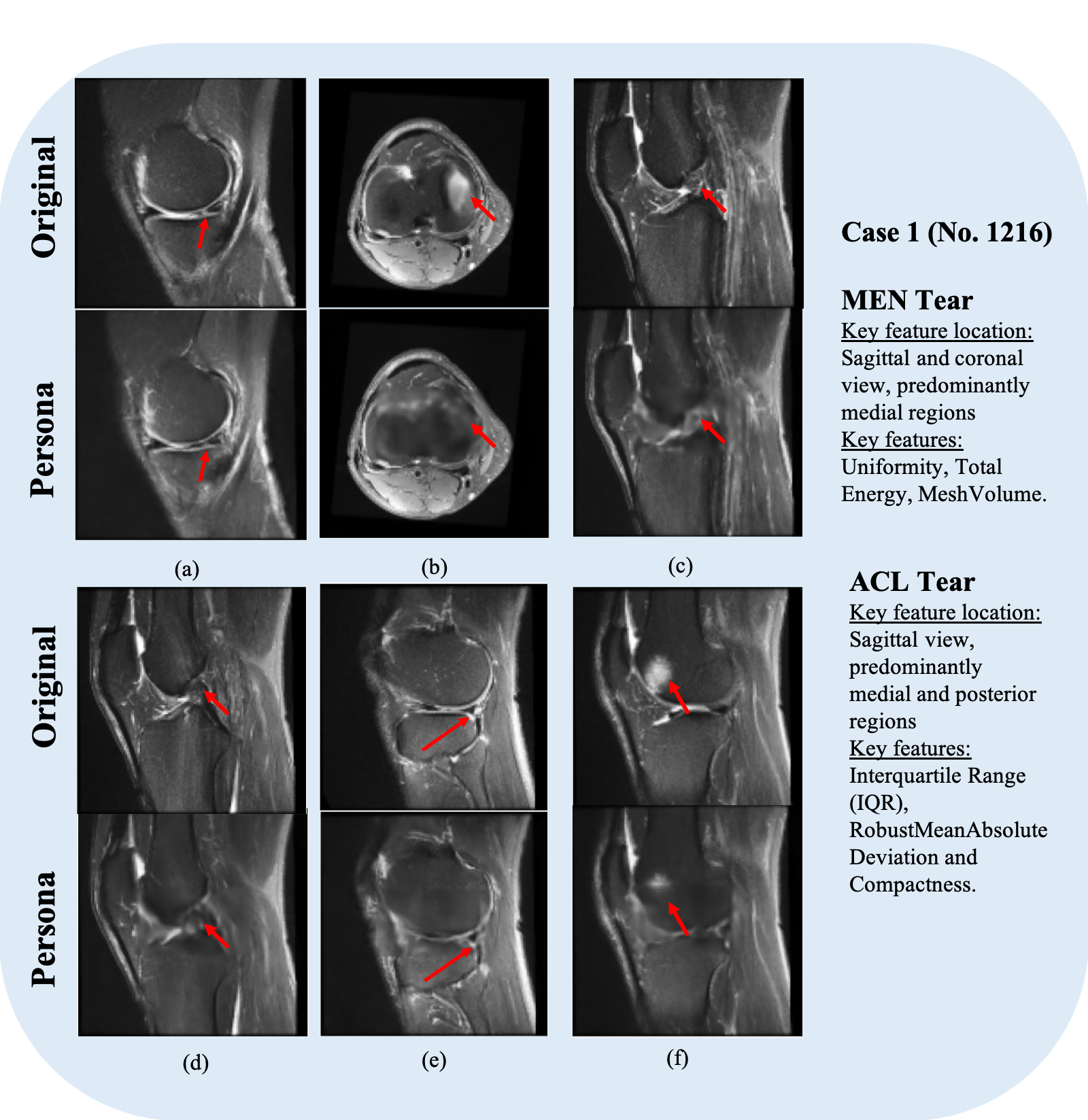}
    \caption{Representative cases of ACL and Meniscus Tears (details in Sec.\ref{case1paragraph}).}
    \label{case1}
\end{figure}

This case presents a complex knee injury involving a medial bucket-handle meniscal tear alongside a partial ACL rupture. The meniscal tear is first identified in Fig.~\ref{case1}a and ~\ref{case1}b. In the original clinical scan (Fig.~\ref{case1}a), a hyperintense signal traverses vertically through both surfaces of the meniscus reaching the joint lines(as indicated by the red arrow). Fig.~\ref{case1}b, an axial view, further illustrates deformation and displacement of the medial meniscus, strongly suggestive of a bucket-handle tear (as indicated by the red arrow). In comparison, the paired persona images show a more regularly contoured meniscus with more uniform signal characteristics, serving as a plausible structural reference to aid in identifying the abnormality in the clinical images. ACL integrity is assessed in sagittal views (Fig.~\ref{case1}c and ~\ref{case1}d). The clinical image reveals partial fiber discontinuity, with residual fibers attached to the tibial insertion, indicative of a partial rupture (as indicated by the red arrow). The persona counterpart shows a more continuous, low-signal ligament, representing a reference morphology that facilitates the determination.

Secondary signs further support this observation. In Fig.~\ref{case1}e, the red arrow points to an area of increased signal intensity in the posterolateral corner (PLC), suggestive of soft tissue injury. The persona image exhibits lower signal in this region, reflecting a more preserved structure. Additionally, bone bruising is evident in the anterior aspect of the lateral femoral condyle (Fig.~\ref{case1}f), characterised by clustered subchondral hyperintensity (as indicated by the red arrow). This finding is notably reduced in the persona, indicating a more physiologic marrow signal pattern.

Radiomic feature importance analysis supports and quantifies these observations. In the case of the meniscal tear, key features are predominantly localised in the sagittal and coronal planes, with focus on the medial patches, consistent with the anatomical location of the tear. Uniformity emerges as a key feature, reflecting decreased signal homogeneity in the torn meniscus, which corresponds to the morphological disruption observed on imaging.

For the ACL tear, key features concentrate in the sagittal view, primarily in the medial and posterior regions, aligned with the anatomical location of the ACL. Interquartile range (IQR) is identified as another key feature, capturing increased local signal variability due to partial fiber disruption and surrounding tissue changes.

\subsection{Case 2: lateral radial meniscal tear with ACL rupture}
\label{case2paragraph}
\begin{figure}
    \centering
     \includegraphics[width=1\textwidth]{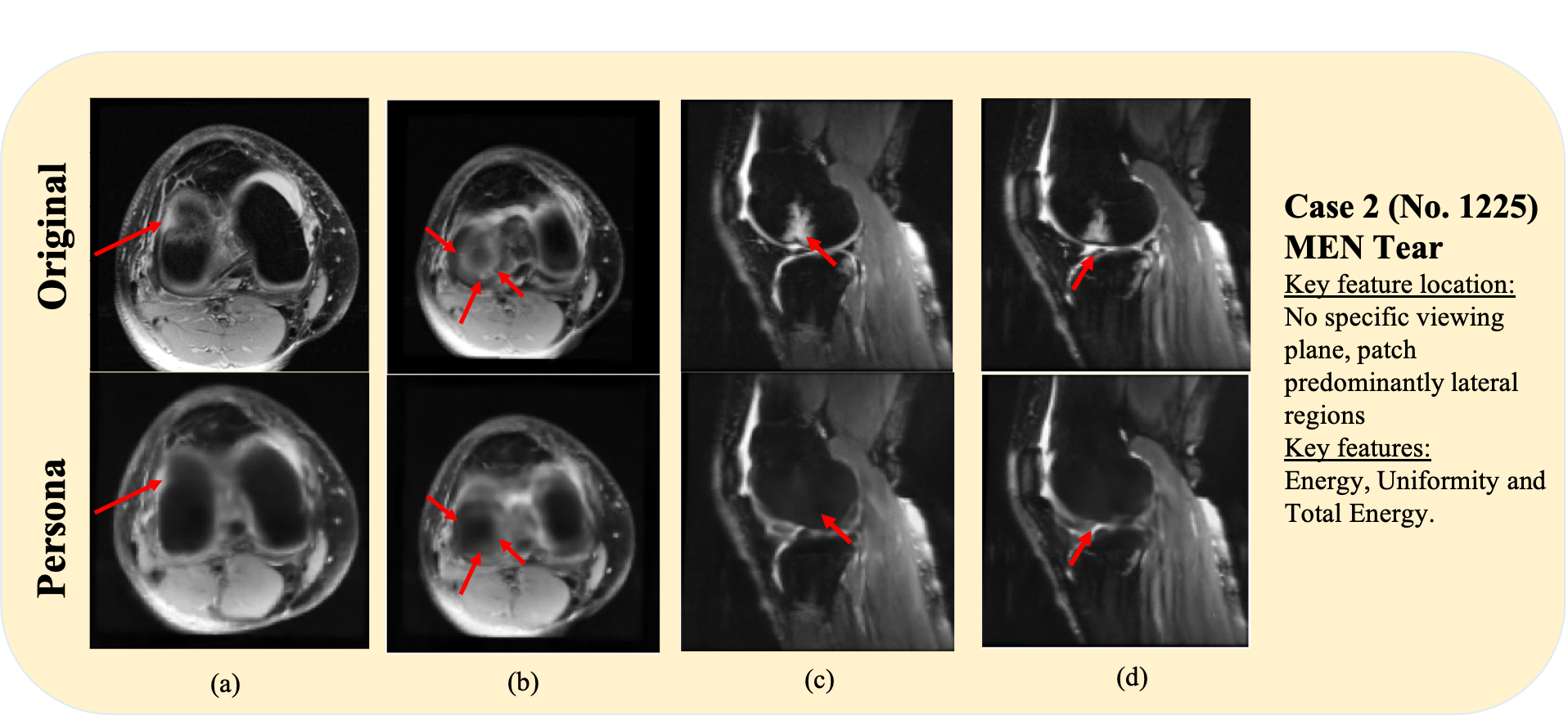}
    \caption{Representative cases of Meniscus Tears (details in Sec.\ref{case2paragraph}).}
    \label{case2}
\end{figure}

This case presents a lateral radial meniscal tear accompanied by an ACL rupture. In the following analysis, we focus primarily on the meniscal pathology. In Fig.~\ref{case2}a, which provides information from the axial plane,  a hyperintense signal is observed in the anterolateral aspect of the lateral meniscus, suggesting possible tearing (as indicated by the red arrow). Fig.~\ref{case2}b presents another piece of evidence: linear hyperintense signals radiate from the central portion of the meniscus toward the periphery, a pattern characteristic of radial tearing. The corresponding persona shows a more homogenous meniscal structure and uniform signal intensity.

Joint effusion and bone bruising are noted in Fig.~\ref{case2}c. The clinical scan shows intra-articular fluid accumulation and subchondral hyperintensity in the lateral femoral condyle, consistent with post-traumatic marrow changes (as indicated by the red arrow). These signs are weakened in the paired persona, which provides a relatively normalized context for visual contrast. Further abnormalities are shown in Fig.~\ref{case2}d, where the clinical image demonstrates the absence of the normal triangular shape of the anterior part of the lateral meniscus (as indicated by the red arrow).

Feature importance analysis for this case suggests that decisive patterns are distributed across multiple planes, with no single dominant view. Notably, the key subpatches contributing to the diagnosis of meniscus tear are primarily located in the lateral patches, likely reflecting the meniscal tear. However, the complexity of this case—including the concurrent ACL rupture and secondary findings such as joint effusion and bone bruising—contributes to signal variation across multiple imaging planes.

Key features include Energy, Total Energy and Uniformity. Energy and Total Energy are radiomic features influenced by signal intensity and image volume, which may reflect fluid accumulation and tissue density changes. Uniformity quantifies the distribution of intensity values, with lower uniformity indicating increased tissue heterogeneity—potentially due to intra-articular fluid or disrupted meniscal structure.

\subsection{Case 3: ACL tear}
\label{case3paragraph}
\begin{figure}
    \centering
     \includegraphics[width=1\textwidth]{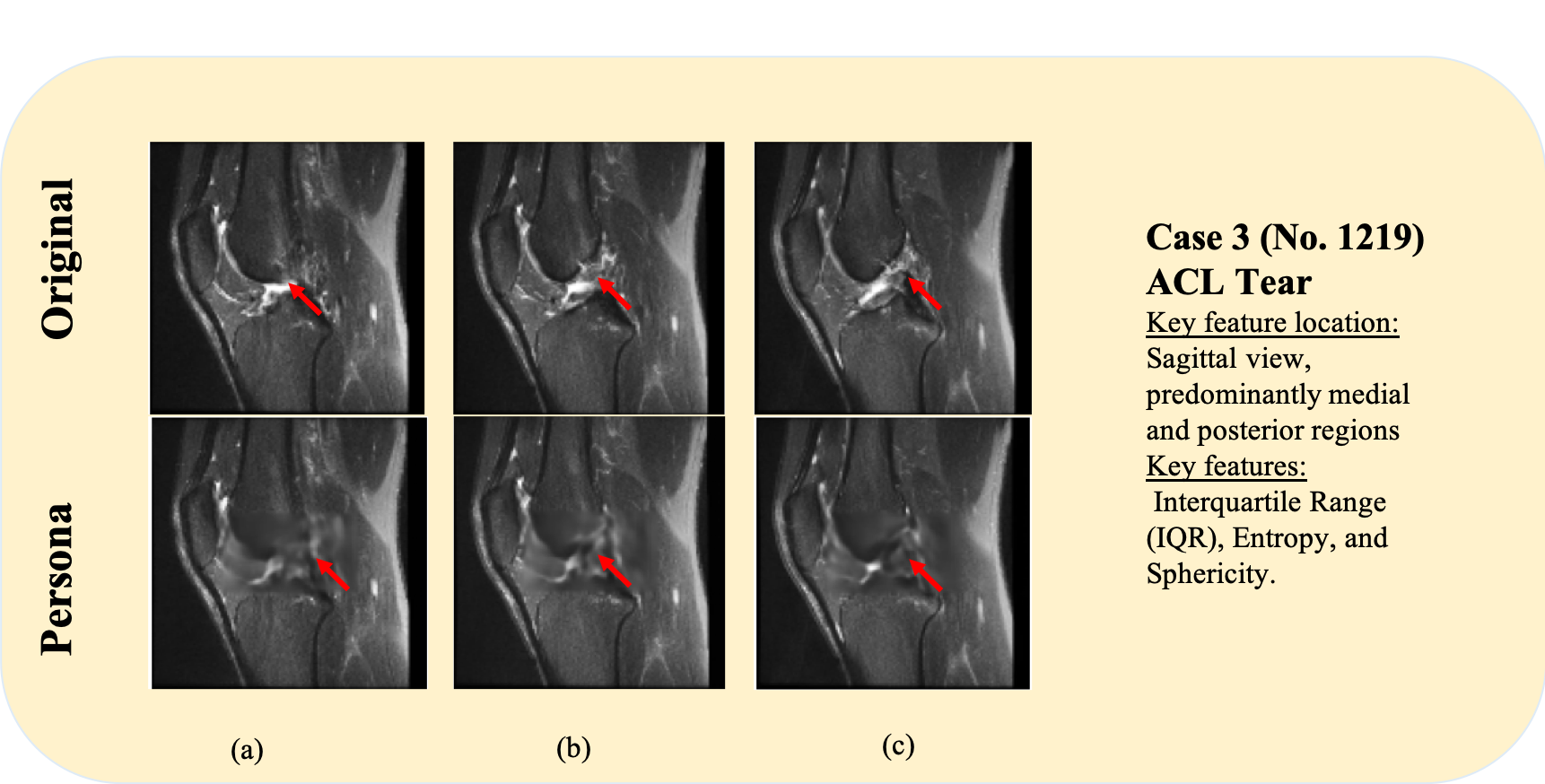}
    \caption{Representative cases of Meniscus Tears (details in Sec.\ref{case3paragraph}).}
    \label{case3}
\end{figure}

This case presents an ACL rupture. The sagittal MRI plane is most suitable for evaluating ACL integrity. Fig.~\ref{case3}a–~\ref{case3}c display the original images, which show clear signs of ligamentous disruption (as indicated by the red arrow). These include fiber discontinuity, abnormal contour and periligamentous hyperintensity. In the corresponding persona images, the ACL appears more continuous and homogeneous in signal. Plus, compared with the original scans, the tibial translation is not clearly evident in the paired persona.

Radiomic feature analysis further supports the diagnosis. Key features in this case are localised primarily in the sagittal view, with dominant contributions from medial and posterior regions, consistent with the anatomical location of the ACL. Among the most informative features are IQR, Entropy and Sphericity. The IQR reflects tissue heterogeneity caused by partial fiber rupture. Entropy quantifies signal complexity and randomness, correlating with the structural disorganization observed in ligamentous injuries. Sphericity, a shape features, may relate to the deformation of the ACL’s normal linear morphology following rupture.

These cases highlight the role of personas in supporting visual interpretability. By providing a reference state, personas can assist in distinguishing both primary and subtle secondary signs of ligamentous pathology, contributing to more informed diagnostic decision-making.

\section{Discussion and conclusion}

Our study presents a novel framework for knee MRI analysis that integrates generative modelling and logistic regression classification to enable an individualized learning-based radiomic feature usage prediction, for pathology classification with improved interpretability. By introducing a patient-specific healthy persona, we provide a baseline comparison that facilitates new insights and clinically relevant analysis of pathological regions. 
Our results demonstrate improvements over conventional end-to-end DL approaches, while maintaining an interpretable feature selection process, is a key advantage of our method. 

\textit{\textbf{\noindent{Patient-specific feature selection balances interpretability and performance.}}} The central contribution of our framework is its ability to adapt feature relevance on a per-patient basis. By weighting radiomic featuress according to the individual anatomy and pathology, the model avoids reliance on a fixed, manually defined feature subset. This patient-specific adaptation improves both diagnostic accuracy and interpretability, particularly in heterogeneous conditions such as ACL and meniscus tears where imaging markers vary considerably across patients.

Our proposed method, built upon first-order intensity statistics and 3D shape featuress, provides a clinically interpretable foundation. These features are widely used in radiology and directly linked to observable image characteristics, making the resulting decisions easier to validate in practice. As confirmed in our ablation study (Table~\ref{tab3}), this configuration achieves strong performance while maintaining transparency.

Extending the feature space to include higher-order texture families (GLCM, GLSZM, GLRLM, NGTDM, GLDM) further improved performance. This suggests that texture radiomics can capture complementary information on intra-tissue heterogeneity and spatial grey-level dependencies that may be overlooked by first-order features. However, such gains came at the expense of reduced interpretability, as higher-order featuress are more abstract and less directly relatable to radiological reasoning.

The addition of the healthy persona further boosted sensitivity (Table~\ref{tab3}). This indicates that the healthy persona amplifies deviations suggestive of pathology. While such behavior may be desirable in screening or triage settings where false negatives are costly, it also risks over-sensitivity to benign variations.

In summary, these findings highlight a fundamental trade-off. The proposed patient-specific selection of first-order and shape features offers an interpretable and robust foundation. Incorporating higher-order features and healthy persona information enhances sensitivity, but at the cost of reduced specificity and interpretability. The optimal configuration may therefore depend on the clinical application—whether the priority is explainable decision support, maximal detection sensitivity, or a calibrated balance between the two.

\textit{\textbf{\noindent{Realism and utility of the healthy persona.}}} The concept of a healthy persona not only augments the radiomic feature pool but also improves the clarity of feature relevance by offering a patient-specific baseline. Our experiments show that including the healthy persona leads to a notable performance boost across all tasks (Table~\ref{tab3}), suggesting that the ability to compare pathological regions against a plausible, disease-free reconstruction is highly valuable for feature discovery and classification. Interestingly, while the quality or “realism” of the reconstructed persona is important, it does not need to be perfect. Rather, it should capture core anatomical plausibility, sufficient to contrast pathological deviations in the radiomic space. We did not observe negative effects such as misleading artifacts from the synthetic persona, but we acknowledge that in settings with more subtle pathologies or low data availability, this risk deserves further study.

\textit{\textbf{\noindent{View- and patch-level personalisation.}}}
Our results suggest that personalised imaging strategies (e.g., sagittal-only for ACL tears) will potentially improve classification performance. These findings are consistent with clinical intuition and radiological practice, where diagnostic information is often concentrated in specific anatomical planes. While we did not implement patch-level personalisation in this study, the observed variability in discriminative subpatch regions across patients indicates that such an approach could further enhance model performance and interpretability. Future work could investigate adaptive spatial decomposition tailored to individual anatomical or pathological presentations, potentially leading to more efficient and patient-specific imaging analyses.

Overall, our framework provides an interpretable, patient-specific radiomic selection mechanism that achieves competitive performance on the state-of the-art DL models, with particularly strong gains for both ACL and meniscus tear classification. Future work will focus on extending personalisation to adaptive view/patch selection and richer radiomic/deep features. The consistent ablation improvements and transparent decision pathway suggest the approach is a promising step towards practical, explainable knee MRI decision support.

\section{Acknowledgement}
This work was supported by the International Alliance for Cancer Early Detection, an alliance between Cancer Research UK [EDDAPA-2024/100014] \& [C73666/A31378], Canary Center at Stanford University, the University of Cambridge, OHSU Knight Cancer Institute, University College London and the University of Manchester; and the National Institute for Health Research University College London Hospitals Biomedical Research Centre.

\section{Disclosure of Interests}
The authors declare that they have no conflicts of interest related to this work.








\bibliographystyle{elsarticle-num} 
\bibliography{reference}

\end{document}